\documentclass[%
 reprint,
 amsmath,amssymb,
 aps,
]{revtex4-2}

\usepackage{graphicx}
\usepackage{dcolumn}
\usepackage{bm}
\usepackage{subfigure}
\usepackage{amsmath}
\usepackage{algorithm}
\usepackage{algpseudocode}
\usepackage{xcolor}
\usepackage{bm}

\begin{document}

\preprint{APS/123-QED}

\title{Do Two AI Scientists Agree?}

\author{Xinghong Fu}
  \email{fxh@mit.edu}
\author{Ziming Liu}%
 \author{Max Tegmark}%
\affiliation{%
 Department of Physics,
  Institute of Artificial Intelligence and Fundamental Interactions,
  Massachusetts Institute of Technology,
  Cambridge, USA
}%

\newcommand{\zm}[1]{{\color{black!0!blue} #1}}

\date{\today}

\begin{abstract}
When two AI models are trained on the same scientific task, do they learn the same theory or two different theories? Throughout the history of science, we have witnessed the rise and fall of theories driven by experimental validation or falsification: many theories may co-exist when experimental data is lacking, but the space of surviving theories becomes more constrained with more experimental data becoming available. We show the same story is true for AI scientists. With increasingly more systems provided in training data, AI scientists tend to converge in the theories they learned, although sometimes they form distinct groups corresponding to different theories. To mechanistically interpret what theories AI scientists learn and quantify their agreement, we propose MASS, Hamiltonian-Lagrangian neural networks as AI Scientists, trained on standard problems in physics, aggregating training results across many seeds simulating the different configurations of AI scientists. \textbf{Our key findings include: 1) when trained on textbook problems in classical mechanics, AI scientists prefers either a complete Hamiltonian or Lagrangian description; 2) when extended to non-standard physical problems, the Lagrangian description generalizes, suggesting that Lagrangian dynamics remain as the singular accurate family of descriptions in a rich theory space.} We also observe strong seed dependence of the training dynamics and final learned weights, controlling the rise and fall of relevant theories. Besides interpretability, MASS unifies and generalizes beyond the Lagrangian neural networks and the Hamiltonian neural networks, providing a new tool for learning of dynamical systems. We release our code at \url{https://github.com/shinfxh/ai-scientists}.
\end{abstract}

\maketitle


\section{\label{sec:level1}Introduction}
\begin{figure}[t]
    \centering
    \includegraphics[width=\linewidth]{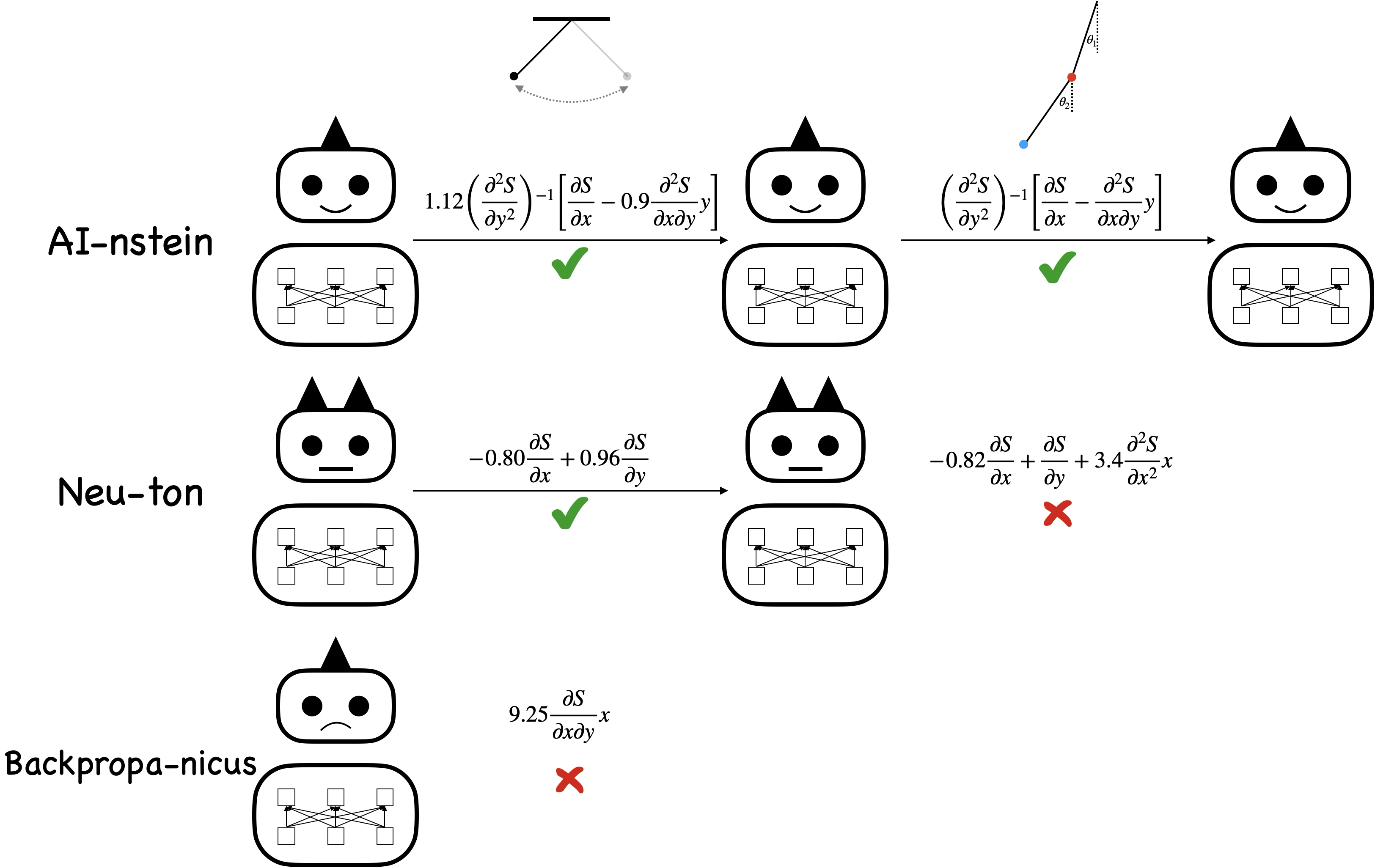}
    \caption{\label{fig:ai_scientists} The evolution of AI scientists. Different AI scientists learning from data within the same physical system, even in the simple pendulum, arrives at different results. Theories that fail to support the current data are marked wrong. Surviving AI scientists are exposed to more complex systems, such as the double pendulum. AI scientists modify their theories to model the new data. Ultimately, what will the remaining AI scientists learn?}
\end{figure}
Throughout human history, our collective curiosity has driven scientific progress. From Archimedes’ principle of buoyancy, to Galileo’s systematic study of motion, to Newton’s formulation of classical mechanics, and finally to Einstein’s revolutionary theories of relativity, these luminaries meticulously analyzed observations and experiments to develop robust hypotheses that explained known phenomena and predicted new ones. Over the centuries, as technology has advanced, so too has our ability to refine experiments, test theories against increasingly precise datasets, and update our frameworks accordingly. Some hypotheses have eventually fallen out of favor, while others have evolved into more nuanced theories capable of describing phenomena at previously uncharted scales \citep{kuhn1970}.

Today, in the twenty-first century, we are witnessing the emergence of a new paradigm. Machine learning (ML) and data-driven methods have already begun to supplant traditional statistical tools in fields as diverse as particle physics \citep{baldi2014searching}, astronomy \citep{ball2010data}, materials science \citep{ramprasad2017machine} and quantum chemistry~\citep{ferminet}. A natural next step is to contemplate a future in which ML methods shift from mere assistive instruments to becoming full-fledged “AI scientists,” capable of formulating hypotheses, designing experiments, and interpreting results with minimal human intervention. Pioneering endeavors have already produced end-to-end AI platforms that can discover physical laws from raw experimental data \citep{schmidt2009distilling, cranmer2020discovering}, and molecular structures from protein sequences \citep{jumper2021highly}. The recent improvements in architectures \cite{attention} with capabilities to absorb and process a large amount of data have fueled the development of large-language models \cite{bert, gpt, llama, mixtral, gemma, deepseek}. These LLMs have already started becoming the backbone of fully-automated AI research scientists \cite{lu2024aiscientistfullyautomated}. 

As these AI scientists begin to operate autonomously, it is worth asking: What scientific theories will they propose? History shows that different human researchers, such as Newton and Leibniz, can arrive at complementary yet distinct formulations of the same phenomenon (e.g., calculus). Analogously, modern ML systems vary in architecture, initialization schemes, and training paradigms \citep{lecun2015deep}, leading to the possibility that independently trained AI scientists might converge on different theoretical formulations or complementary perspectives. Moreover, as AI scientists venture into analyzing larger and more intricate datasets—ranging from high-dimensional cosmological surveys to complex molecular dynamics simulations—their learned representations and theories may evolve in unexpected ways \citep{carleo2019machine}.

This paper does not seek to predict precisely how AI will transform science in the decades ahead. Instead, we offer a set of controlled experiments to investigate whether and how multiple AI scientists, trained under varying conditions, converge or diverge in their scientific theories. By exploring synthetic datasets, we aim to shed light on how the complexity of the data, the choice of model architectures, and the selection of training methods may influence not only what these AI systems learn but also how their internal representations and resulting theories develop over time \citep{rudin2019stop}.

In doing so, we hope to provide a window into the kind of questions that will shape future inquiries into the role of AI in science: will AI scientists unify disparate theories or fragment into multiple, equally valid viewpoints? Will their theories be understandable to humans, or will interpretability become a bigger challenge? The experimental framework and preliminary results presented herein serve as a stepping stone for these discussions, highlighting the potential—and potential pitfalls—of our emerging AI Scientists.

To list, our contributions in this paper are: 
\begin{enumerate}
    \item We propose a new architecture, MASS (Multiple AI Scalar Scientists), to allow a single neural network to learn diverse theories across multiple physical systems.
    \item We train MASS across datasets including the simple pendulum, the Kepler problem, and synthetic potentials.
    \item We analyze significant activations in MASS and distill the theories learned by MASS.
\end{enumerate}

Using MASS as a proxy for an AI scientist, our findings suggest that 

\begin{enumerate}
    \item An AI scientist can learn many diverse explanations of the same physical phenomenon.
    \item Encountering more complex systems, successful AI scientists modify their existing theories to suit new observations.
    \item AI scientists tend to learn similar theories, evaluated in terms of similarity of the networks' internal activations. These theories also closely resemble either the Hamiltonian or Lagrangian description.
    \item The recovered theories resemble Hamiltonian dynamics initially, then shift closer to a Lagrangian formulation as complexity of systems increase. This suggests that the Lagrangian formulation remains as the only correct theory within a rich theory space.
\end{enumerate}

\section{Related Work}


Scientists aim to recover equations from observation. So do AI scientists. Given a set of data of some physical system, we aim to uncover the underlying ``truth" in terms of physical equations. Efforts to tackle this problem has been a mix of discrete methods such as combinatorial optimization, making use of methods involving genetic programming~\citep{koza1994genetic} and continuous ones centering around symbolic regression~\citep{schmidt2009distilling}. The underlying assumption is that there are few number of terms in the final expression, inspiring methods of sparse linear regression~\citep{sindy}. Physical priors were introduced~\citep{aifeynman} to improve the ability of symbolic regression techniques in discovering known physical equations. In this paper, we propose a method to discover underlying physical laws with minimal physical priors, making use of the principle of stationary action, learning a single scalar function. These two properties are shared by the Hamiltonian Neural Network (HNN) \cite{hnn} and the Lagrangian Neural Network (LNN) \cite{lnn}.


Inspired by the Hamiltonian formulation of classical mechanics, The HNN breaks down the task of learning the equations of motion of a physical system to first learning a scalar function, the Hamiltonian $H$, then obtaining $(\dot{q},\dot{p})$ using  Hamilton's canonical equations: 
\begin{equation} \label{eqn:ham}
\dot{q} = \frac{\partial H}{\partial p}, \quad \dot{p} = - \frac{\partial H}{\partial q},
\end{equation} 
where $q, p$ are the canonical position and momentum. However, in some cases the expression for these canonical coordinates is not trivial to write down. The LNN solves this problem by learning the scalar function as the Lagrangian instead and taking derivatives according to the Euler-Lagrange equations. 

\begin{equation} \label{eqn:lag}
\frac{d}{dt}\left(\frac{\partial \mathcal{L}}{\partial \dot{q}}\right) - \frac{\partial \mathcal{L}}{\partial q} = 0. 
\end{equation}
This avoids the need for an explicit expression of canonical momentum, making LNNs advantageous for certain physical systems.

Since the introduction of these works, there has been significant leaps in improving the efficiency of training~\citep{generalizedlnn, constraintlnn}, as well as applying these networks to problems in domains such as rigid-body dynamics~\citep{lnn_rigid_body}, particle interactions~\citep{lnn_particle}, video prediction~\citep{lnn_video} and generative modeling~\citep{hnn_generative}. However, in many of these works the underlying equations of motion (Equations \ref{eqn:ham} and \ref{eqn:lag}) are baked into the model architecture, and the model resultingly learns the corresponding theory governed by this equation. Instead, we ask the following question: when the model is given a freedom of learning multiple theories, what will it learn?

In this work, our proposed model, \emph{Multiple AI Scalar Scientists} (MASS), is a generalized framework that includes both LNN and HNN as special cases. MASS is similarly inspired by the principle of stationary action. Like LNNs and HNNs, we aim to learn a free-form scalar function from data. However, unlike LNN and HNNs, which have hard-coded equations of motion, we equip MASS with the ability to also learn the equations of motion. For a physical system described by generalized coordinates $q$ and velocities $\dot{q}$, one can learn a scalar function (akin to a Lagrangian or Hamiltonian) that governs the system’s evolution, such that the path obeys the principle of stationary action. 



The architecture design of MASS allows it to learn a rich space of theories, defined by the coefficients on each term in the governing equation learned by MASS. Similar to \citep{lnn}, our experiments are done in generalized coordinates. Through a series of controlled experiments on an ensemble of MASS scientists in these coordinates, we will probe the underlying theories learned.

\section{MASS: The AI scientist} \label{sec:mass}
\begin{figure}[t]
    \centering
    \includegraphics[width=\linewidth]{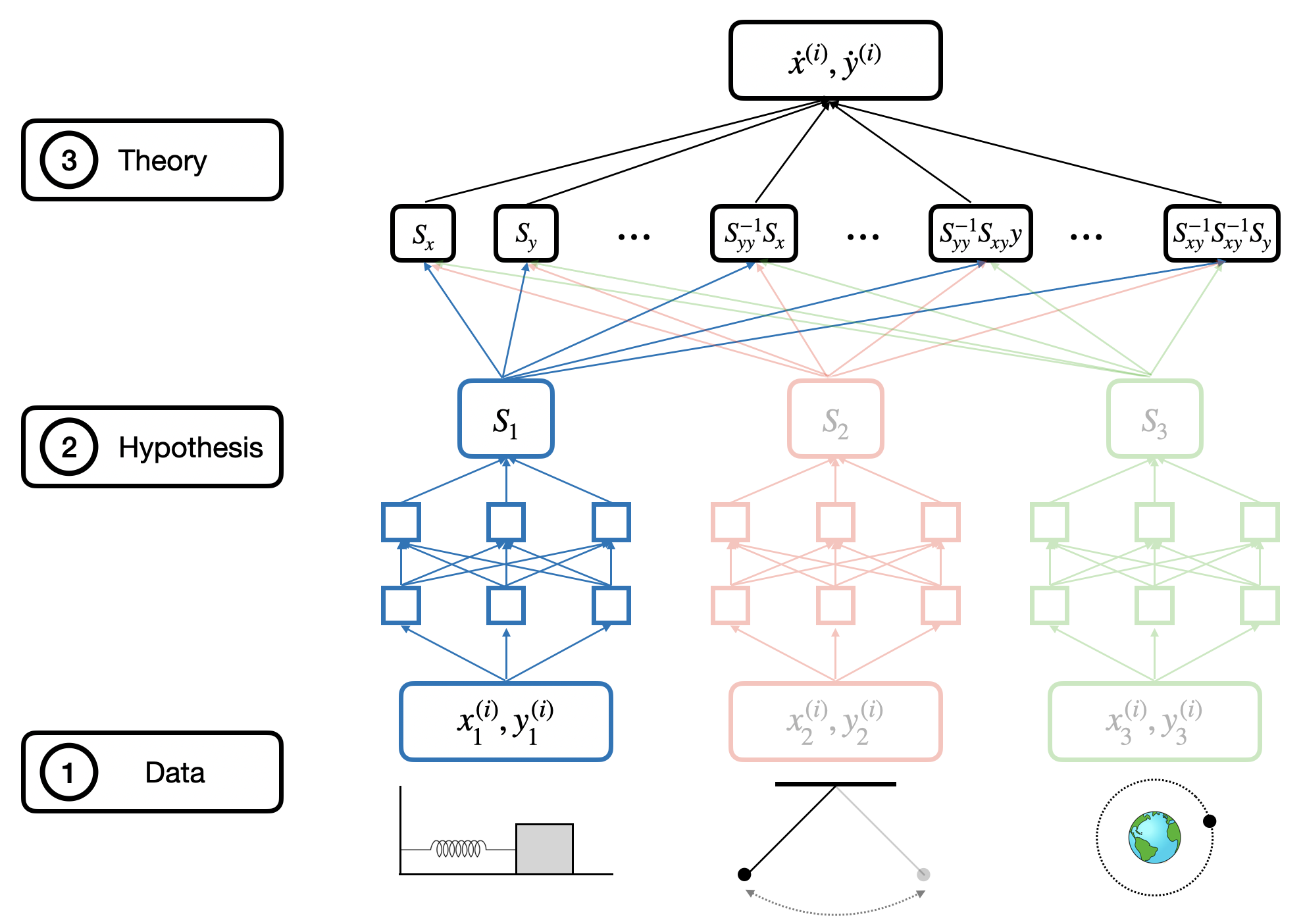}
    \caption{\label{fig:mass} The MASS (Multi-physics AI Scalar Scientist) network. }
\end{figure}

To emulate how human scientists operate, the core idea behind MASS is to embed within a single neural network the capacity to learn and unify information from multiple physical systems. Rather than fitting individual models for each system, MASS aims to internalize a \emph{shared} framework that captures fundamental patterns across all datasets. Specifically, it achieves this by learning a \textbf{scalar function}---analogous to Lagrangian or Hamiltonian---whose derivatives encode the system-specific dynamics. As illustrated in Figure~\ref{fig:mass}, MASS adopts the following workflow:

\begin{enumerate}
    \item \textbf{Data ingestion:} MASS receives observational data (e.g., trajectories, states, or energy values) from diverse physical systems such as pendulums, orbital problems, or other synthetic potentials.
    \item \textbf{Hypothesis formation:} A separate neural network for each system learns a single scalar function that encapsulates the system-specific dynamics.
    \item \textbf{Theory evaluation:} A final layer shared across all systems differentiates the learned scalar function with respect to the system coordinates (position, momentum, and/or velocity), MASS infers a system’s governing equations. This enforces the consistency of an overarching theory across multiple systems.
    \item \textbf{Refinement and generalization:} The outputs of the model, in this case the time derivatives of the input, are then evaluated against the ground truth training data to compute errors. The error is summed across all systems, then backpropagation optimizes a single theory that is simultaneously consistent with multi-physics observations.
\end{enumerate}

By iterating through these steps, MASS strives to discover a single, scalar function for each system and a shared final layer forming a generalized theory across systems. Together, the scalar function and the weights in the final layer (i.e. how MASS takes its derivatives) form the \textit{theory} that it learns.

\section{Method} \label{sec:method}

We denote by $\mathcal{M}$ a single MASS scientist network. $\mathcal{M}$ learns from $n$ different physical systems. Some examples of systems include the spring-mass system, the gravitational system, and quantum mechanical systems, etc. Each system obeys some underlying physical law, be it the inverse square law of gravitational attraction, or Schr\"odinger's equation. For simplicity and as a proof of concept, we constrain our systems to classical mechanics below. 

\textbf{Data ingestion}: The input variables for system $j$ to $\mathcal{M}$ are the generalized coordinates in $d$-dimensions, expressed as $\textbf{x}_j, \textbf{y}_j \in \mathbb{R}^d$, where $\textbf{x}_j$ and $\textbf{y}_j$ are the generalized coordinates and their time derivative, respectively. For a simple pendulum, we can either express $(\textbf{x}_j, \textbf{y}_j) := (\theta, \dot{\theta})$ as a one-dimensional problem, or with $\textbf{x}_j := (x, y)$ and $\textbf{y}_j := (\dot{x}, \dot{y})$ to express the problem in two dimensions with Cartesian coordinates.


\textbf{Hypothesis formulation}: This block consists of of $n$ separate neural networks each learning a separate potential function $S_j$ for each system $j$.  We denote this forward pass as 
\begin{align}
    S_j = f_j(\textbf{x}_j, \textbf{y}_j).
\end{align}
In this paper, we focus on MLPs, which suffice for learning $S$.

\textbf{Theory evaluation}: The shared derivatives layer computes the derivatives, up to second-order, of $S_j$ with respect to the input variables $\textbf{x}_j,\textbf{y}_j$. Note that given $d$-dimensional inputs, i.e. $\textbf{x}_j,\textbf{y}_j \in \mathbb{R}^d$, the single-variable derivatives $S_\textbf{x}, S_\textbf{y} \in \mathbb{R}^d$ are column vectors while the second-order derivatives (and their inverses) are Hessian matrices, i.e. $S_\textbf{xx}, S_\textbf{yy}, S_\textbf{xy}, S_\textbf{xx}^{-1}, S_\textbf{yy}^{-1}, S_\textbf{xy}^{-1} \in \mathbb{R}^{d \times d}$. To allow the network to learn a diverse set of \textit{theories}, we compute all products of terms, up to three terms in the product, such that the final result is a $\mathbb{R}^d$ vector that predicts the time derivatives $\dot{\textbf{x}}_j, \dot{\textbf{y}}_j \in \mathbb{R}^d$. Particularly, let the set of $\mathbb{R}^d$ vectors be $\mathcal{V} = \{\textbf{x}, \textbf{y}, S_\textbf{x}, S_\textbf{y}\}$ and the set of $\mathbb{R}^{d \times d}$ matrices be $\mathcal{A} = \{S_\textbf{xx}, S_\textbf{yy}, S_\textbf{xy}, S_\textbf{xx}^{-1}, S_\textbf{yy}^{-1}, S_\textbf{xy}^{-1}\}$. there are three types of terms that can potentially predict $\dot{\textbf{x}}_j, \dot{\textbf{y}}_j$: 
\begin{enumerate}
    \item $\textbf{v} \in V$
    \item $\textbf{A} \textbf{v}$ where $\mathbf{A} \in \mathcal{A}$ and $\textbf{v} \in V$
    \item $\textbf{A}_1 \textbf{A}_2 \textbf{v}$ where $\textbf{A}_1, \textbf{A}_2 \in \mathcal{A}$ and $\textbf{v} \in V$.
\end{enumerate}
In our implementation, there are a total of $T=172$ different terms across the three types described above, and we explicitly compute them using
\begin{align}
    \textbf{t}_j = D(f_j(\textbf{x}_j, \textbf{y}_j)).
\end{align}
where $D$ is the derivative layer and $\textbf{t} \in \mathbb{R}^{T \times d}$ gives the terms that can potentially enter into the final equation. 

In the final layer, the network learns a linear combination of these $\mathbb{R}^d$ vectors to predict the time derivatives of the inputs. Note that since we are using generalized position and momentum, $\dot{\textbf{x}} = \textbf{y}$ trivially up to a constant factor. The remaining of the paper focuses on investigating the set of activations in the final layer that predicts $\dot{\textbf{y}}$. We denote this final layer as $L_f$, and the output prediction of $\dot{\textbf{y}}$ will be given by 
\begin{align}
    \hat{\dot{\textbf{y}_j}} = L_f(\textbf{t}_j)=L_f(D(f_j(\textbf{x}_j, \textbf{y}_j))).
\end{align}

\textbf{Refinement and generalization}: For a specific system $j$, we predict $\hat{\dot{\textbf{y}_j}}$ and then compute the MSE loss with the ground truth data. We then sum up the losses across all systems that $\mathcal{M}$ is learning over, and do a backpropagation on the accumulated gradients. After convergence, the model develops a theory that is consistent across its multiple physical systems. The optimization objective is written as 
\begin{align}
    \min_\theta \sum_{j = 1}^{n} \mathbb{E}_{(\textbf{X,Y})} ||\dot{\textbf{Y}}_j - \hat{\dot{\textbf{Y}_j}}||_2^2,
\end{align}
where $\textbf{Y}_j \in \mathbb{R}^{N \times d}$ is the concatenation of $N$ samples in system $j$, and the expectation is taken over the samples $\textbf{X, Y}$ drawn i.i.d. for each system. 

We find that optimization over $\theta$, the parameterization of $\mathcal{M}$ is highly unstable (as observed in~\cite{lnn}), due to the computation of derivatives and inverses in the matrices $\mathcal{A}$. The experimental procedure and hyperparameter setting is more detailed in Appendix \ref{appendix:hyper}, but some key design choices help achieve stable training: 
\begin{itemize}
    \item Computing the pseudo-inverse with a regularized stabilization term. Instead of computing ${\rm inv}(S_\textbf{xx})$, we compute ${\rm pinv}(S_\textbf{xx} + b)$ where $b$ is penalized as a regularization term in training. 
    \item AdamW \cite{adamw} optimizer with cosine learning rate scheduling \cite{cosine} and warm restarts.
    \item Augmenting inputs to include second-order terms of $\textbf{x, y}$.  
\end{itemize}

\section{Experiments}

\subsection{Single scientist: Correlated theories} \label{sec:1s1s}
\begin{quote}
``It might be that to describe the universe, we have to employ different theories in different situations. Each theory may have its own version of reality, but according to model-dependent realism, that is acceptable so long as the theories agree in their predictions whenever they overlap, that is, whenever they can both be applied."
    \\
    \hfill --- Stephen Hawking \& Leonard Mlodinow, \textit{The Grand Design} (2010)
\end{quote}
\begin{figure}[ht!]
    \centering
    \subfigure[Training dynamics]{
        \includegraphics[width=0.2\textwidth]{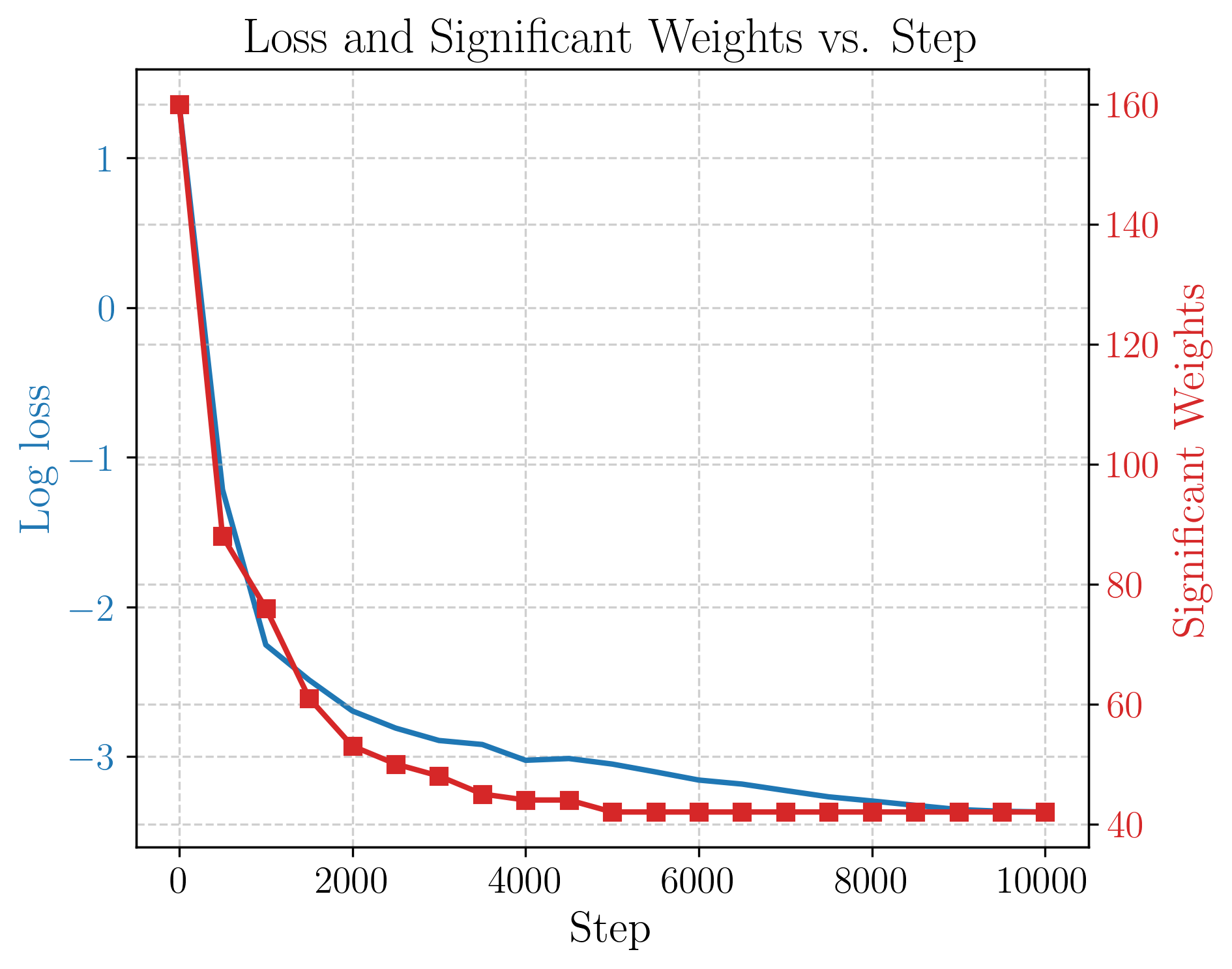}
    }
    \subfigure[Simulated trajectory]{
        \includegraphics[width=0.2\textwidth]{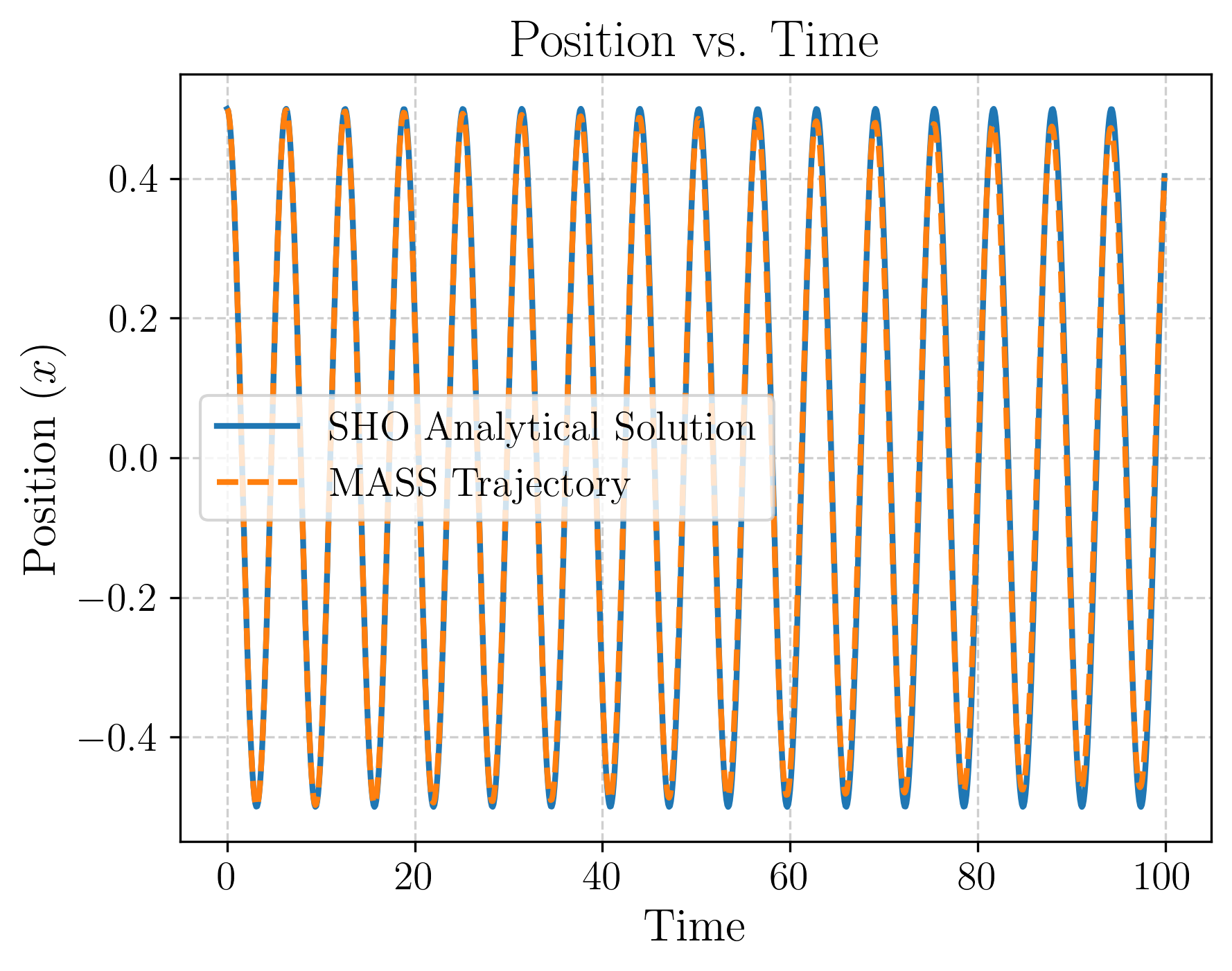}
    }
    \caption{Training results for MASS on the simple harmonic oscillator. \textbf{(a)} MASS (seed 0) trains to an MSE loss of $3 \times 10^{-4}$ over 10000 steps of a batch size of $512$ at each step. The number of significant weights, calculated as the number of weights in the final layer that account the first 99\% the total norm, decreases with loss. \textbf{(b)} The recreated motion of a single oscillator accurately captures the frequency and amplitude of the motion. }
    \label{fig:1s1s_training}
\end{figure}

The central message of Hawking’s statement is that multiple theoretical frameworks can provide equally valid descriptions of physical phenomena, as long as their predictions agree with experiment. A typical classroom demonstration of this principle is the undamped spring-mass system. One can appeal to Newton’s laws of motion, where the governing equation is 
\[
    m \ddot{x} = - k x,
\]
or switch to a Hamiltonian formulation in which energy functions and conservation laws offer an alternative viewpoint.

Machine learning models, on the other hand, tend to be overparameterized, often giving them considerable flexibility in fitting data, even for relatively simple physical systems~\cite{zhang_rethinking,allen_overparam}. An intriguing question arises: 
\emph{If we train a single ``AI scientist'' network on a simple harmonic oscillator, what sort of theoretical representation will it learn, and how will it compare to the standard Newtonian or Hamiltonian descriptions?}

To investigate this, we trained MASS on simulated data from the undamped spring-mass system. Figure~\ref{fig:1s1s_training} shows the training results. Particularly, we observe that training on the simple harmonic oscillator is not a difficult task for MASS, converging to an MSE loss of $3 \times 10^{-4}$. We are interested in understanding how the model learns and simplifies its theory, under addition of $L_1$ and $L_2$ regularization to the final layer. To that end, we track the number of significant weights across training steps, calculated as the number of weights in the final layer that account for the first $99\%$ total norm of the final layer weight vector. Observe that this also decreases with the total number of training steps, but it plateaus at a rather large number of 42. This means that almost $42$ weight terms have significant magnitude, not at all close to a simple theory of $\dot{y} = -x$!

Using MASS, we can also simulate a trajectory of the oscillator rather easily, and Figure~\ref{fig:1s1s_training} show the consistency of predictions given by MASS.

\begin{figure}[ht!]
    \centering
    \subfigure[Contour Plot]{
        \includegraphics[width=0.2\textwidth]{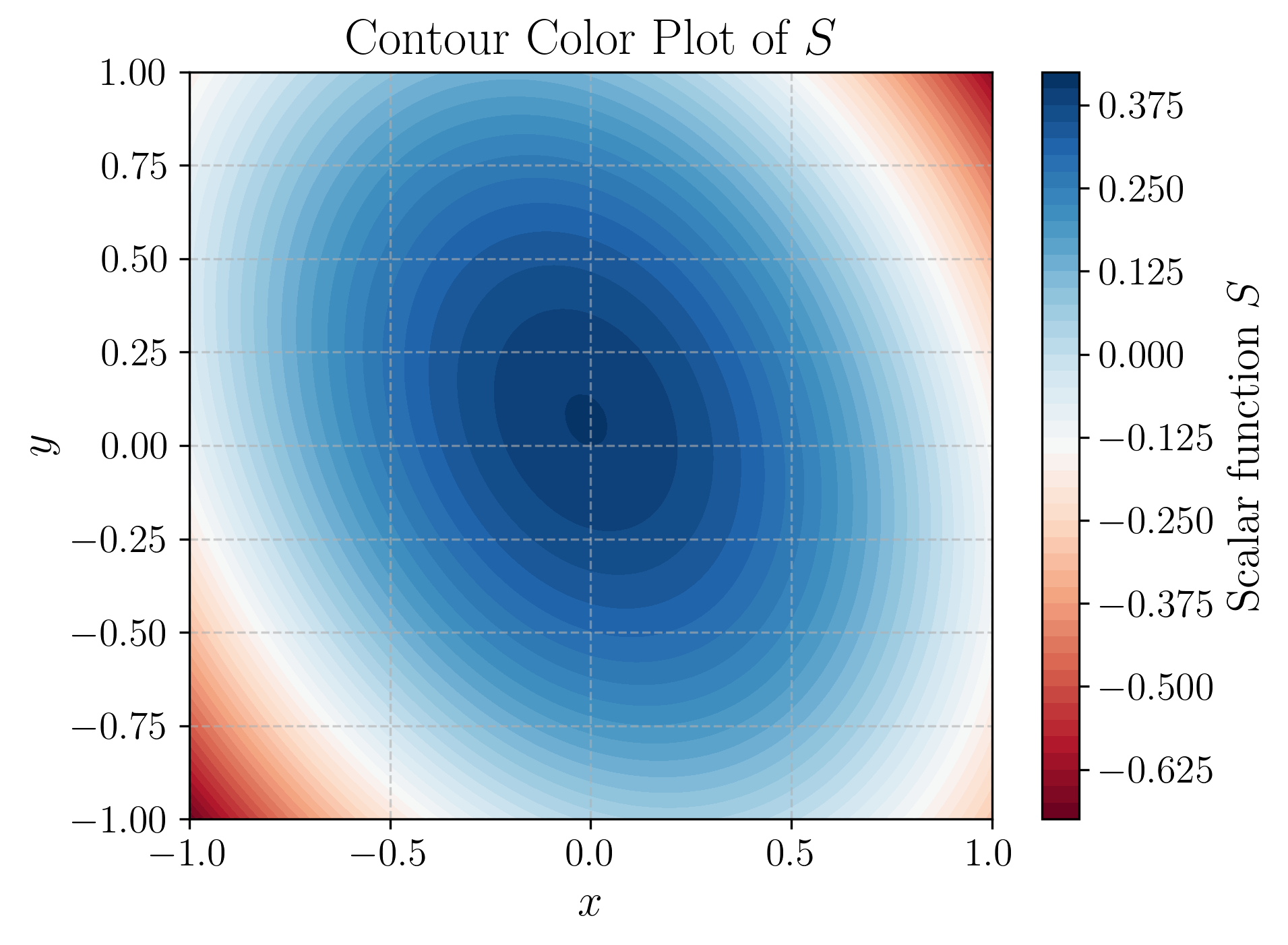}
    }
    \subfigure[Hamiltonian $x^2+y^2$]{
        \includegraphics[width=0.2\textwidth]{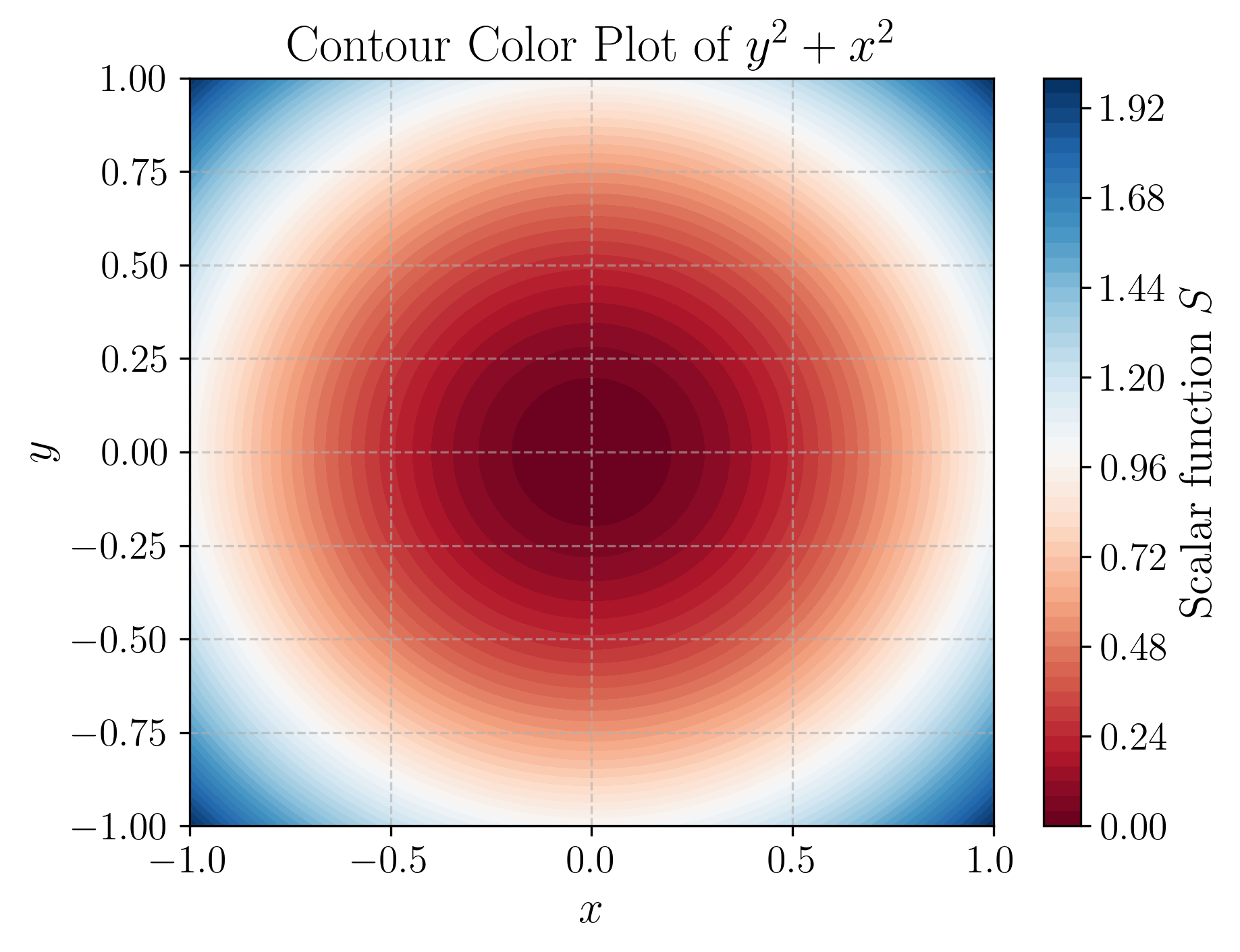}
    }
    \caption{Contour of \textbf{(a)} learnt scalar function $S$, compared with \textbf{(b)} the Hamiltonian $x^2+y^2$. MASS can in general, learn functions that resemble yet differ from conventional physical priors.}
    \label{fig:contour_weights_31}
\end{figure}

Figure~\ref{fig:contour_weights_31} depicts the learned scalar function \( S \) over phase space, compared to a canonical Hamiltonian function $H$. A single MASS scientist is able to recover the sum of potential and kinetic energy expression. However, we already see some differences between these two expressions. To note, our learnt $S$ is convex rather than concave as in $H$. $S$ also takes on negative values, not typically allowed in expressions of energy. In this example and also in general, the contour can also look skewed, translated or even resemble an entirely different conic section. This goes to show the richness of the theory space offered to MASS. 

While the weights $w_i$ in the final layer offer a glimpse as to which terms are important, they are not the full story. Firstly, the contribution to the final prediction of $\dot{y}$ exists as the sum 
\begin{align*}
    \hat{\dot{y}} = \sum_{i = 1}^{T} w_i d_i
\end{align*}
where $d_i$ $i$-th derivative term, such as $S_x, S_{xy}, \dots$

\begin{figure}[ht!]
    \centering
    \subfigure[Weights and Activations]{
        \includegraphics[width=0.2\textwidth]{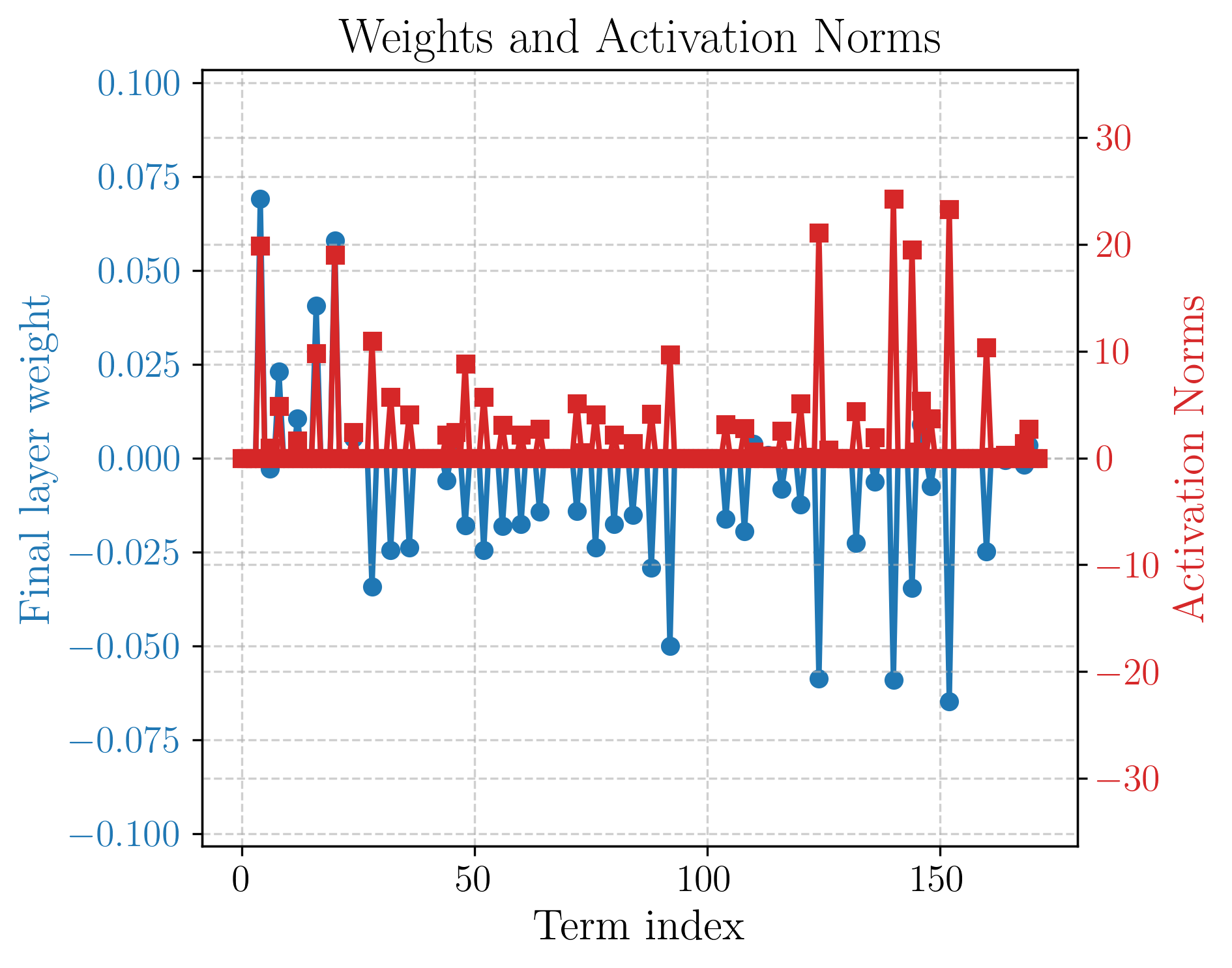}
    }
    \subfigure[Correlation map (Activations)]{
        \includegraphics[width=0.2\textwidth]{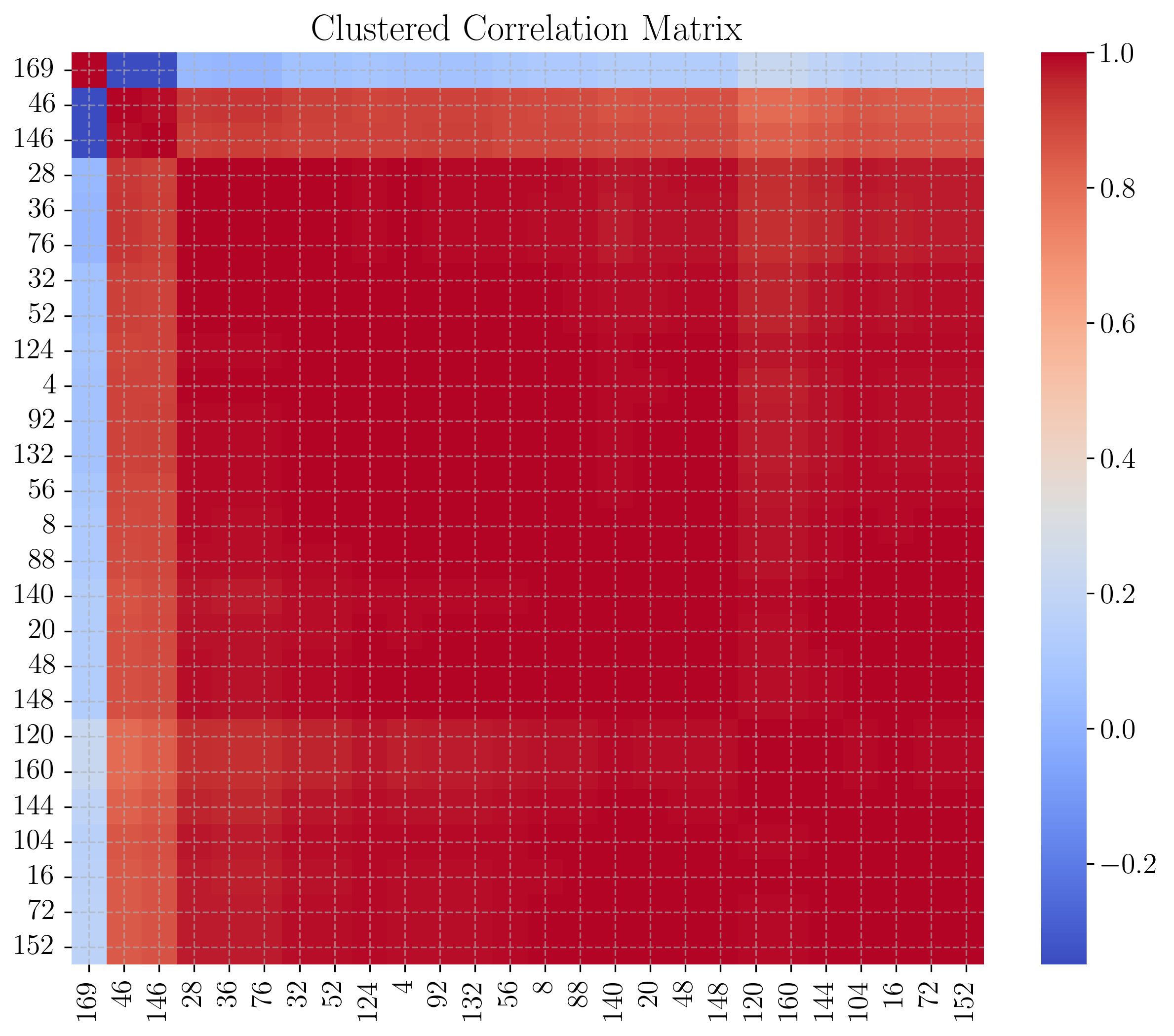}
    }
    \caption{\textbf{(a)} Weights in final layer (blue) and the mean activation norms (red). The top 5 terms in mean activation magnitude: $S_{yy}^{-1} S_{yy}^{-1} x, \; S_{xy}^{-1} S_{yy} x, \; S_{yy}^{-1} S_{xx} x, \; S_{xx} x, \; S_{yy}^{-1} S_{xy}^{-1} x$. \textbf{(b)} Correlation of significant activations, keeping only indices $i$ contributing to the first cumulative $99\%$ of $\sum_i \mathbb{E}[a_i]$, plotted after hierachial clustering. Most terms are strongly correlated.}
    \label{fig:1s1s_corr}
\end{figure}

We hence compute the activation vector $a_i = w_i d_i$ over a sample batch of 512 data points. In Figure \ref{fig:1s1s_corr}, we compare the mean norm of the activations $\mathbb{E}(a_i)$ (expectation taken over the 512 data points) with the weights, $w_i$. In general, non-zero weights correspond to non-zero activation norms, but the relative order of the magnitudes of each term are not necessarily preserved. Particularly, the inverse of Hessian matrices, such as $S_{xx}^{-1}$ are large when the second derivatives of $S$ is small. 

The largest magnitude activations that contribute to the final prediction of $\dot{y}$ are in descending order: $S_{xx} x, S_{xy}^{-1} S_{yy}x, S_{yy}^{-1}S_{yy}^{-1} x, S_{yy}^{-1} S_{xx}x, S_{yy}^{-1} x, \dots$. When sorting instead by the norm of the weights, the top 5 terms are $S_{xx} x, \; S_{xy}^{-1} S_{yy} x, \; S_{yy}^{-1} S_{yy}^{-1} x, \; S_{yy}^{-1} S_{xx} x, \; S_{yy}^{-1} x$. The similarity of these terms is strong indicator of the important terms contributing to the final theory learnt by MASS. 

In the next step, we filter to the number of significant terms. Following the convention in Figure~\ref{fig:1s1s_training}, we keep only those terms that contribute to the first 99\% of the total magnitude. That is, $j$ is the number of significant terms if $\sum_{i=1}^j\mathbb{E}[|a_i|] > \sum_{i=1}^T \mathbb{E}[|a_i|]$. On these remaining significant activations, we compute the correlations in the heatmap in Figure~\ref{fig:1s1s_corr}, sorted according to similarity in correlations by hierarchical clustering \cite{sneath1973numerical}. There are three distinct clusters, consisting of terms that are linear matrix products of the vectors $y, S_x$ and $x$ respectively, from the upper left to lower right. 

The existence of multiple terms lies in contrast to a trivial theory in which the only significant activation is $-x$, that gives a perfect prediction of $\dot{y}$. The Hamiltonian expression would construct $S = \frac{1}{2} x^2 + \frac{1}{2} y^2$ and predict $\dot{y} = -S_x$. The reason behind the multi-expressivity of the network is that most second order terms are constant when the scalar function $S$ is at most second order. For instance, one can easily conceive a network that learns $S = x^2 + y^2 + xy$, where Hessian matrices $S_{xx}, S_{yy}, S_{xy}$ and their inverses become constant products. The invariance in learning these products give rise to the mix of expressions we see. Nonetheless, these terms turn out to be extremely correlated and they mainly represent only one theory. In the next section, we will discuss how the significant terms evolve, which terms survive and which die, when the AI scientist is exposed to more complex physical systems.

The \textbf{main takeaways} from this section is 
\begin{enumerate}
    \item A single AI scientist can very effectively learn a single simple system (Figure~\ref{fig:1s1s_training}), and it learns to filter its theory as training progresses.
    \item The underlying theory resembles some familiar physical function (Figure~\ref{fig:contour_weights_31}).
    \item When incorporated with large capacities, a single AI scientists tends to learn many seemingly separate theories (Figure \ref{fig:1s1s_corr}(a)).
    \item However, many of these theories are strongly correlated (Figure \ref{fig:1s1s_corr}(b)).
    
\end{enumerate}

\subsection{Multiple systems: Sparsification and diversification} 
\label{sec:1s5s}
\begin{quote}
    ``In the beginning of the twentieth century,
    it became apparent that the motion of the planet Mercury
    was not exactly right. This caused a lot of trouble and was
    not explained until it was shown by Einstein that Newton’s
    Laws were slightly off and that they had to be modified. ''
    \\
    \hfill --- Richard Feynman, \textit{The Character of Physics Law} (1967)\citep{Feynman1967}
\end{quote}
The simple harmonic oscillator is perhaps too simple for a machine learning model to fit. You see, it just has to fit $-x$. We now extend our experiments to investigate what happens when an AI scientist, when starting out by observing just a single system, encounters more complex physical systems. Following the training paradigm in Section~\ref{sec:mass}, MASS learns a separate scalar function for each system, while sharing the same final layer. We aggregate the loss across all the systems in one step of training. The specific systems of interest here:
\begin{enumerate}
    \item Simple harmonic oscillator
    \item Simple pendulum
    \item Kepler (Gravitational potential)
    \item Relativistic harmonic oscillator
\end{enumerate}

\begin{figure}[ht]
    \centering
    \includegraphics[width=\linewidth]{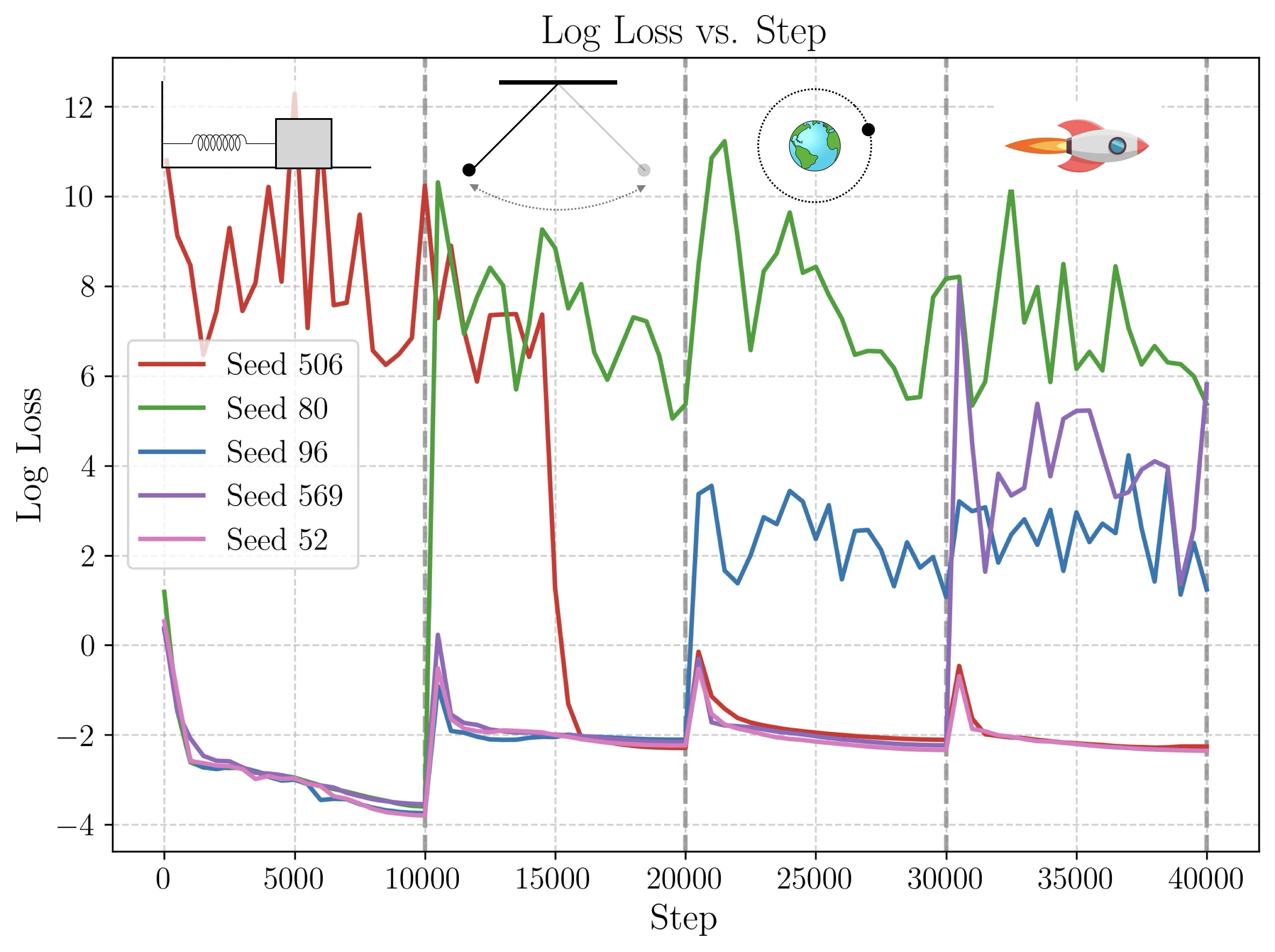}
    \caption{\label{fig:1sms_training} MASS trained on increasingly complex systems. The dashed lines indicate the different phases of training. Starting from the \textit{simple harmonic oscillator}, the system is exposed to the \textit{simple pendulum, gravitational potential} and \textit{relativistic harmonic oscillator} at the $10000^{th}, 20000^{th}, 30000^{th}$ step respectively. Loss is aggregated over all systems MASS is exposed to at each step of training.}
\end{figure}


Figure~\ref{fig:1sms_training} shows the training results as we introduce each system one after another at intervals of 10000 steps, in the aforementioned order, i.e. a single \textbf{training phase} lasts for 10000 steps. This specific order represents at a high level the level of complexity of the systems to a human scientist. We observe that as more systems are introduced, existing theories either survive or falter, depending on the random seed controlling the initialization of the MASS network. For example, seed 80 fails at the simple pendulum, seed 96 fails at the gravitational potential, and seed 569 fails at the relativistic potential. This means that although they survived previous tasks, they probably discovered ``wrong'' theories that only overfit to previous tasks. It is also interesting to note that while some MASS fails initially, they can start learning accurate representations when tasked with a more complex system. The intuition for this late start can be understood by the larger number of constraints that more complex systems impose on MASS, that help its convergence. 

The aggregated behaviors across seeds at each system will be further discussed in Section~\ref{sec:ms1s}. In the remaining of this section, we analyze a single MASS and its surviving terms. Particularly, this will be seed 52.

\begin{figure}[ht!]
    \centering
    \includegraphics[width=\linewidth]{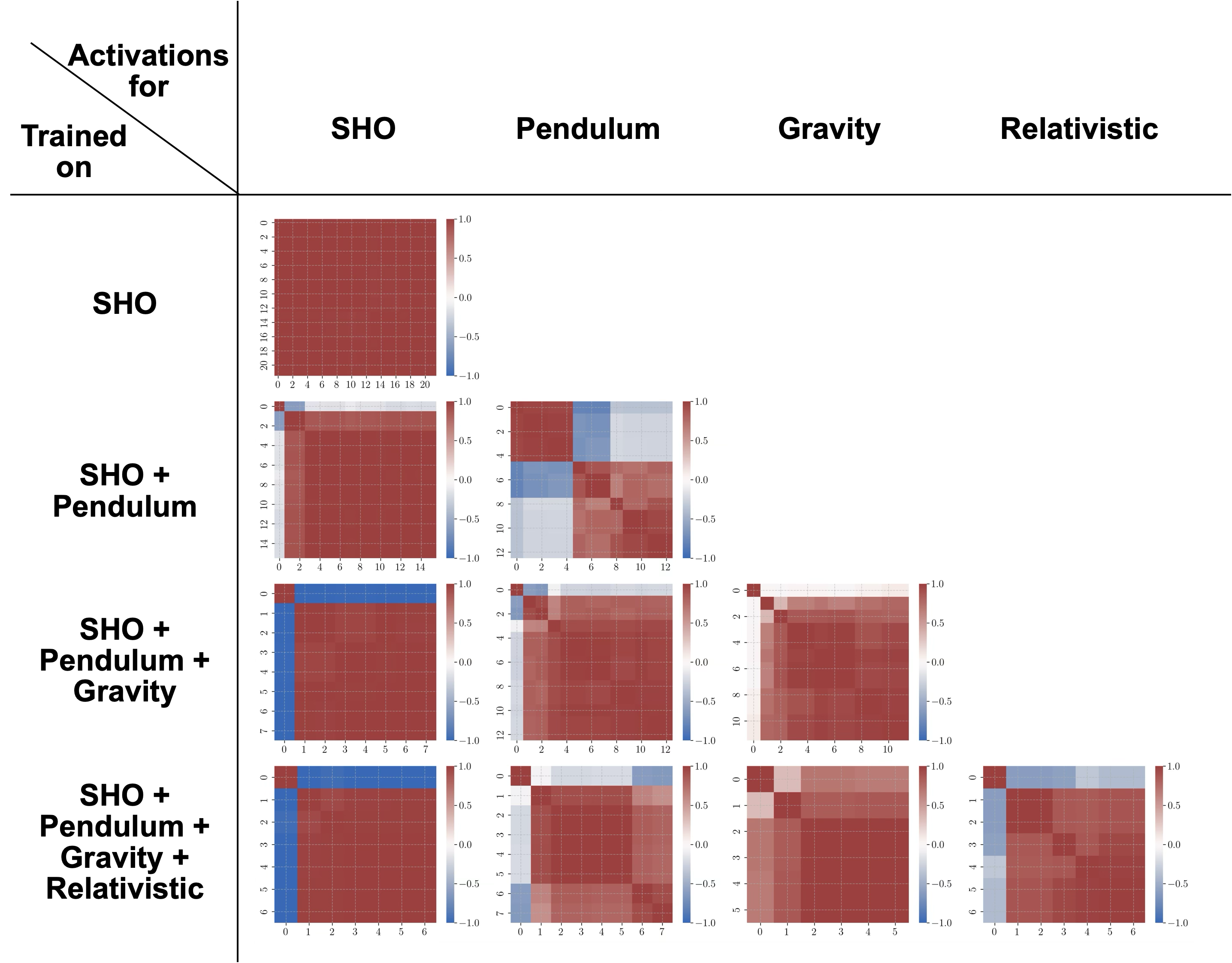}
    \caption{\label{fig:1sms_corr} (Seed 52) Pairwise correlations learnt by a single AI scientist trained on increasingly complex systems moving down. From the top, each rows correspond to activations at steps 10000, 20000, 30000 and 40000 respectively. Correlation map plotted after filtering for significant terms, which contribute the first $99\%$ of the total magnitude of the activation vector. The number of significant terms, shown by the number of distinct squares, decrease as the number of systems increase. Correlation map plotted after hierarchical clustering. Note that many of these terms are strongly correlated (either positively or negatively).}
\end{figure}

Similar to section \ref{sec:1s1s}, we again analyze the activations in Figure~\ref{fig:1sms_corr}. \\

In general, we make the following observations:
\begin{enumerate}
    \item As the number of systems increases, the number of distinct terms learned decreases.
    \item As the number of systems increases, the theories become more diverse.
\end{enumerate}

The first result, from Figure \ref{fig:1sms_corr} that the number of significant terms, counted by the number of squares in each correlation map of Figure \ref{fig:1sms_corr}, decrease from 20 to 6 for the SHO, 12 to 7 for the pendulum and 10 to 5 for the gravitational problem, show that fewer terms can simultaneously explain all the systems, as opposed to just a smaller subset of the systems. The second result is observed from the increasing occurrence of non-correlated terms trending towards the bottom right of Figure~\ref{fig:1sms_corr}. We also find that when tasked with explaining an ensemble of systems, MASS uses almost the same terms! To see this, the last row of Figure \ref{fig:1sms_corr} comprises essentially the same 6 to 7 terms used for explaining all 4 systems. These terms correspond to 
\begin{align*}
    S_{yy}^{-1} x, \; S_{xx} S_{xx} x, \; S_{xx}^{-1} S_{xx}^{-1} x, \; \\
     S_{yy} S_{yy} x, \; S_{xy} S_{xx}^{-1} x, \; \\
     S_{xx}^{-1} S_{yy} x, \; S_{xx}^{-1} S_{xy} x, \; \\
     S_{yy} S_{xx}^{-1} x, \; S_{xx}^{-1} S_{yy} x.
\end{align*}
The large dependence on $x$ is hypothesized to survive from the initial explanation of the simple harmonic oscillator, in which MASS begins by learning terms that are constant products of $x$, and from these terms it develops a theory for the new systems. We verified in a separate set of experiments, that permuting the order of the systems, starting with more difficult systems first, results in more emphasis on terms more related with $S_x$ and $y$. 

We state succinctly the main conclusion from this section: \textbf{as an AI scientist encounters more systems, the number of distinct terms decrease}.

\subsection{Multiple scientists: Mixture of theories}\label{sec:ms1s} 
\begin{quote}
  ``For a brief period at the beginning of 1926, it looked as though there were, suddenly, two self-contained but quite distinct systems of explanation extant: matrix mechanics and wave mechanics. But Schrödinger himself soon demonstrated their complete equivalence.''
  \\
  \hfill --- Max Born, \textit{Nobel Prize Lecture} (1954)
\end{quote}
When multiple scientists work on the same problem independently, some arrive at theories that seem vastly different but later become obvious that they were just two sides of the same coin (think Newton and Leibniz's description of calculus). Differences in theory, reconciled later, happen more so in today's advances in machine learning \cite{sohl2015deep, du2019implicit, song2019generative, song2020scorebased}. Whereas, in some other instances, theories remain different from each other, though they both obey the same experimental results, very much like the Hamiltonian and Lagrangian scalar function descriptions of classical mechanics.

In this section, we investigate the relations between the theories learnt by different MASS scientists (which we will represent by different initial seeds) studying the same system. 

\begin{figure}[ht!]
    \centering
    \subfigure{
        \includegraphics[width=0.22\textwidth]{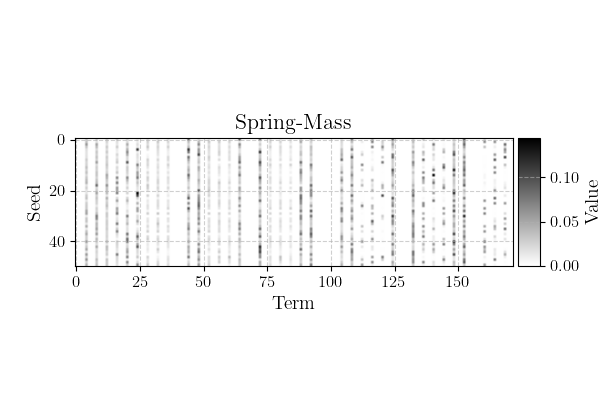}
    }
    \subfigure{
        \includegraphics[width=0.22\textwidth]{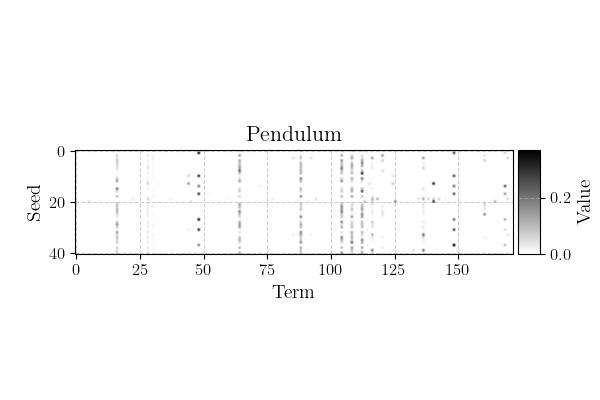}
    }
    \subfigure{
        \includegraphics[width=0.22\textwidth]{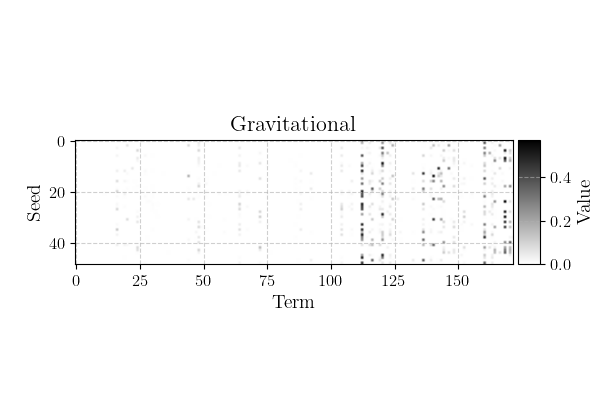}
    }
    \subfigure{
        \includegraphics[width=0.22\textwidth]{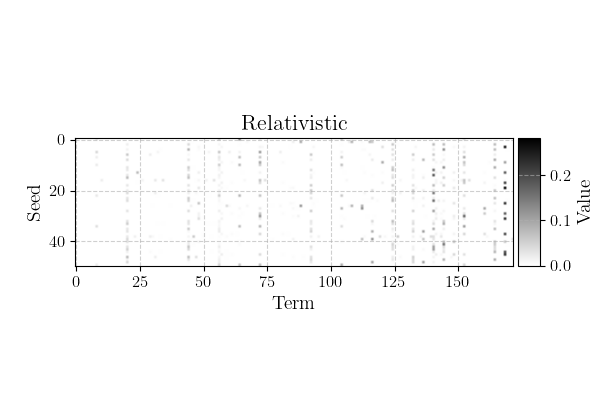}
    }
    \caption{Activation strengths in 50 MASS scientists studying various physical systems separately. Darkness represents stronger activations. The distinct vertical lines indicate the terms on which it is possible, under the MASS framework, to learn a theory of the underlying system.}
    \label{fig:strips}
\end{figure}

The exact weights and values of each activation differs a lot between different scientists. Depending on the initialization, the exact terms which matter changes drastically (refer to Figure \ref{fig:multi_scientists_classical_acts} and more in Appendix \ref{sec:1d_acts}). While the magnitudes of the individual terms vary, the significant terms chosen by each scientist remains rather identical. We illustrate the relative magnitude of each activation term in Figure \ref{fig:strips}. Observe that there are clear lines along this strip, indicating the terms on which it is \textit{possible} to learn a describing theory of the system. 

Nonetheless, the large variations in activation magnitudes and weights indicate that while the theories learned by MASS all lie within the dark lines in Figure \ref{fig:strips}, it might very well be the case that each scientist learns something different. Examining the scalar functions $S$ learnt by individual AI scientists(refer to Figure \ref{fig:many_contours} in Appendix \ref{appendix:scalar}), it is difficult to tell the underlying similarities and differences. Are these AI scientists all learning something entirely different. We will now show that this is not the case. 

Consider the activations on the final layer of MASS which has a shape $B \times T$ on a batch of $B$ samples where the final layer has $T$ terms. Specifically in our case, we have $B = 512, T = 172$. We conduct dimensionality reduction by PCA. It turns out that in majority of seeds, the first principal component already explains more than 90\% of the variance. Reducing into this first principal component gives the $B \times 1$ set of activations, and we observe in Figure \ref{fig:multi_scientists_classical_hist} that in fact, each of the activation values are in fact distributionally equivalent to a uniform distribution (see Figure ~\ref{fig:multi_scientists_classical_hist}).

\begin{figure}[ht!]
    \centering
    \subfigure[SHO]{
        \includegraphics[width=0.22\textwidth]{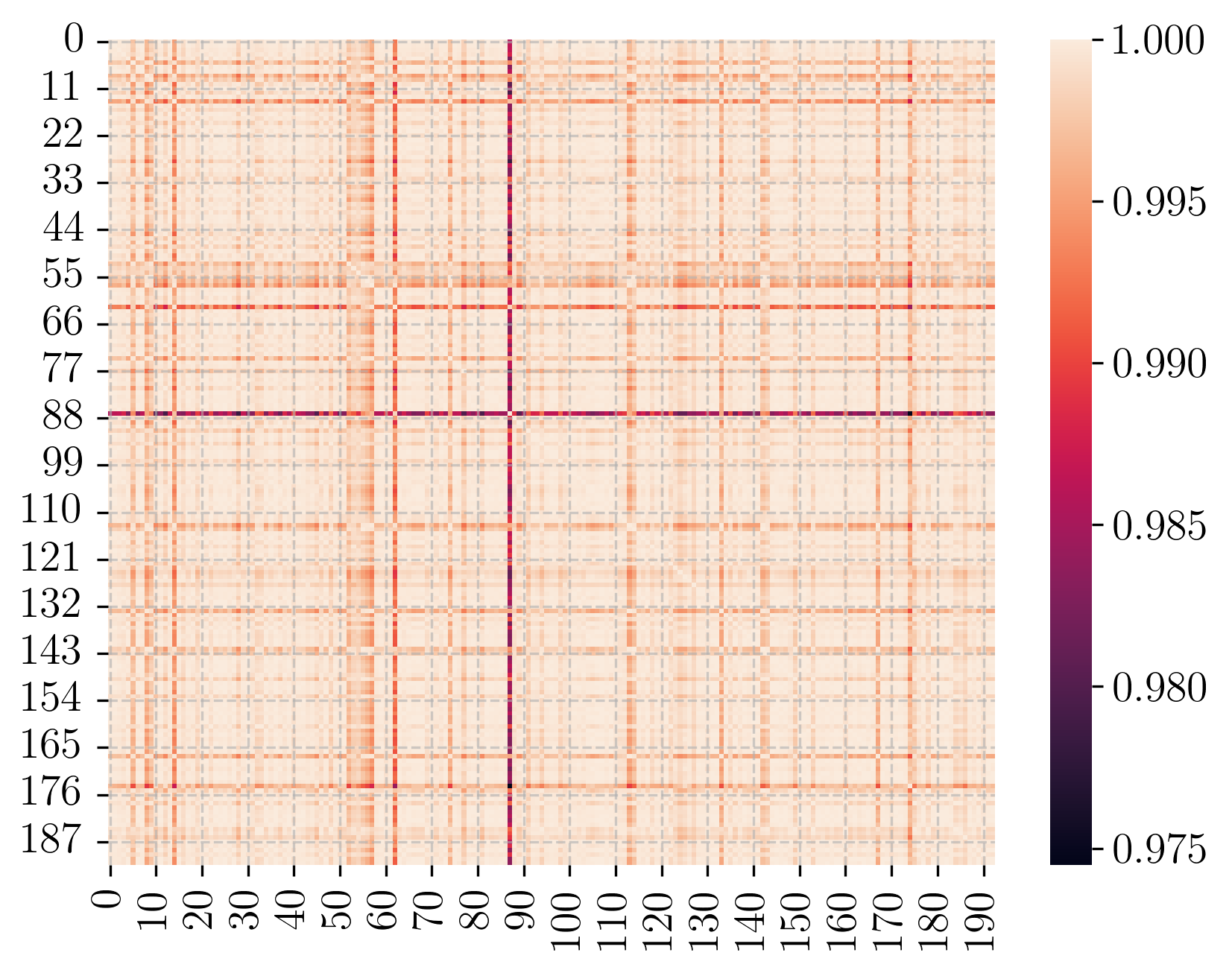}
    }
    \subfigure[Pendulum]{
        \includegraphics[width=0.22\textwidth]{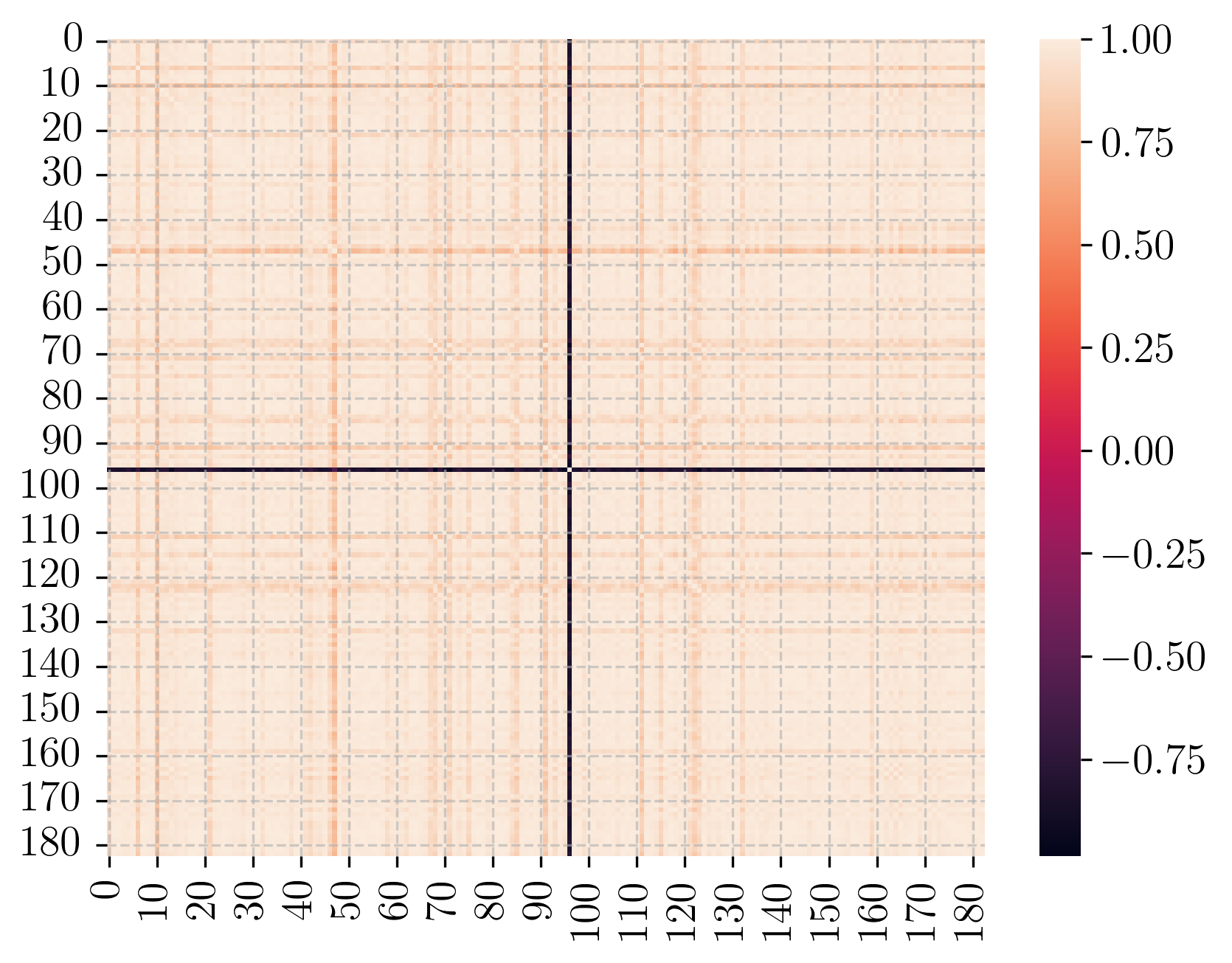}
    }
    \subfigure[Gravitational]{
        \includegraphics[width=0.22\textwidth]{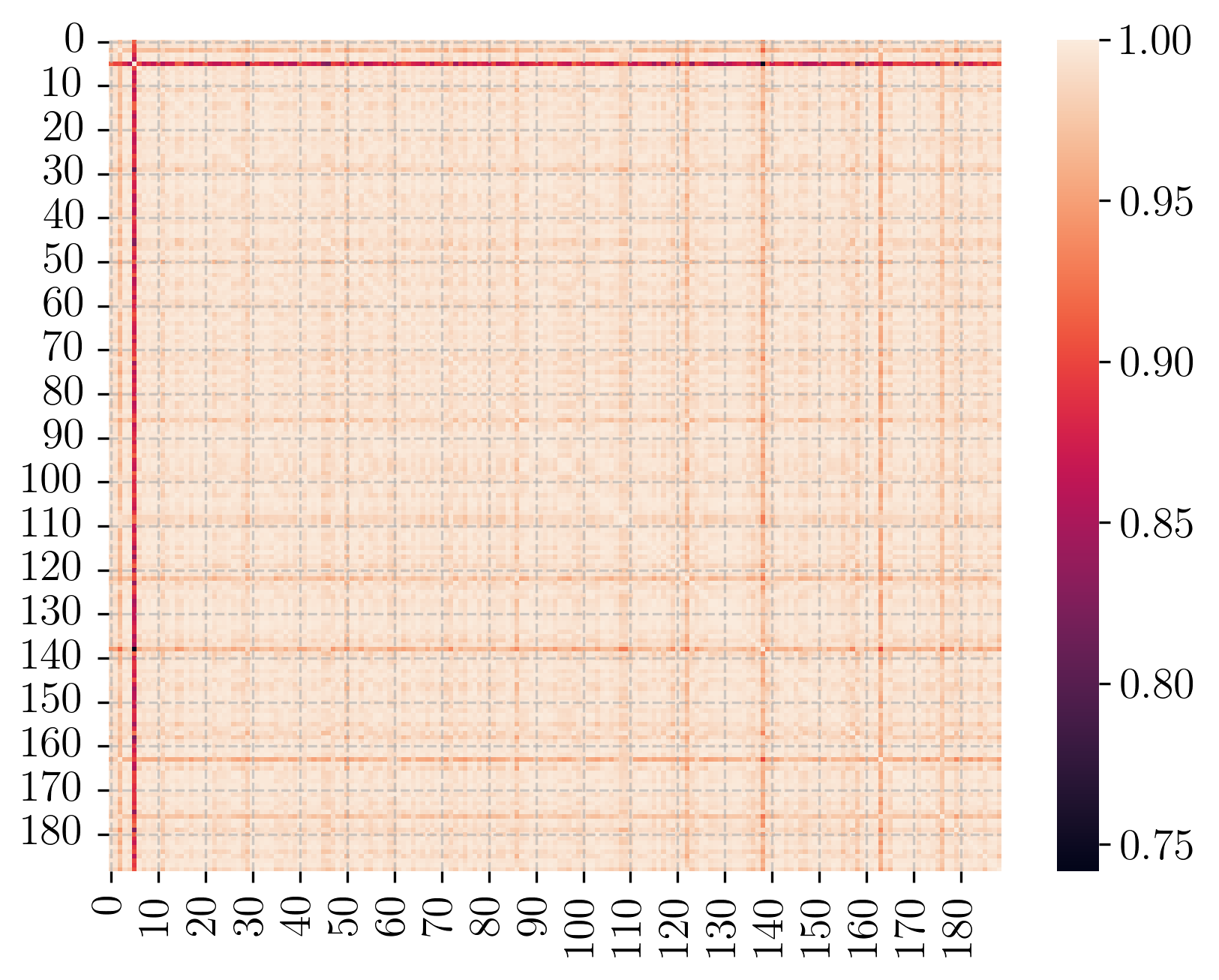}
    }
    \subfigure[Relativistic]{
        \includegraphics[width=0.22\textwidth]{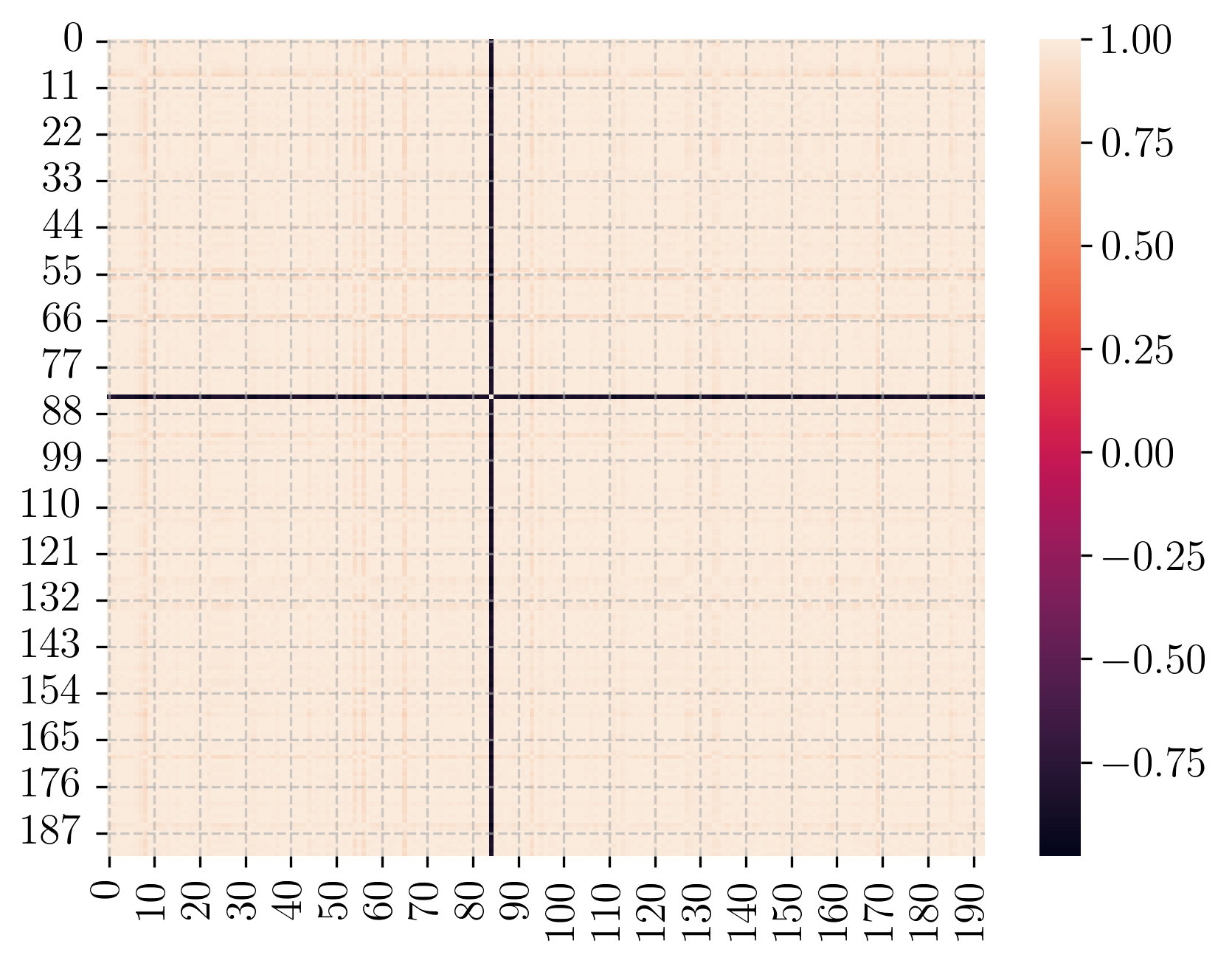}
    }
    \caption{Correlations of the first principal component in 50 MASS scientists studying various physical systems separately. Majority of correlations are high, with the exception of correlations close to $-1$ representing a parity flip. 96.4\%,  74.8\%, 93.7\%, 87.5\% of seeds have their first PCA component explaining more than 80\% of variance for systems \textbf{(a), (b), (c), (d)} respectively.}
    \label{fig:ms1s_corr}
\end{figure}

Such observations are corroborated across multi-scientists set-ups when run on the relativistic spring-mass and the simple pendulum, as given in Figures \ref{fig:multi_scientists_pendulum_hist} and \ref{fig:multi_scientists_relativistic_hist}. 

Computing the correlations between the $B \times 1$ activations shows that each scientist is in fact strongly correlated with all others (see figure~\ref{fig:ms1s_corr}). Note that correlations close to $-1$ denote parity flips, which is surprisingly only rarely learnt. 

These results allow us to conclude that \textbf{multiple scientists learn the same underlying theory when trained on the same physical system}. In fact, this already gives the answer to our very first research question: \textbf{two AI scientists do agree}! 

\subsection{Exploring the unknown: Lagrangian is all you need}
\begin{quote}
    ``I think that the prize is recognizing, in part, the fact that understanding the deep problems of things like mind is not going to come forth in some simple way like Newtonian physics. ''
    \\
    \hfill --- John Hopfield, \textit{Nobel Prize Interview} (2024)
\end{quote}

\begin{figure}[h!]
    \centering
    \includegraphics[width=\linewidth]{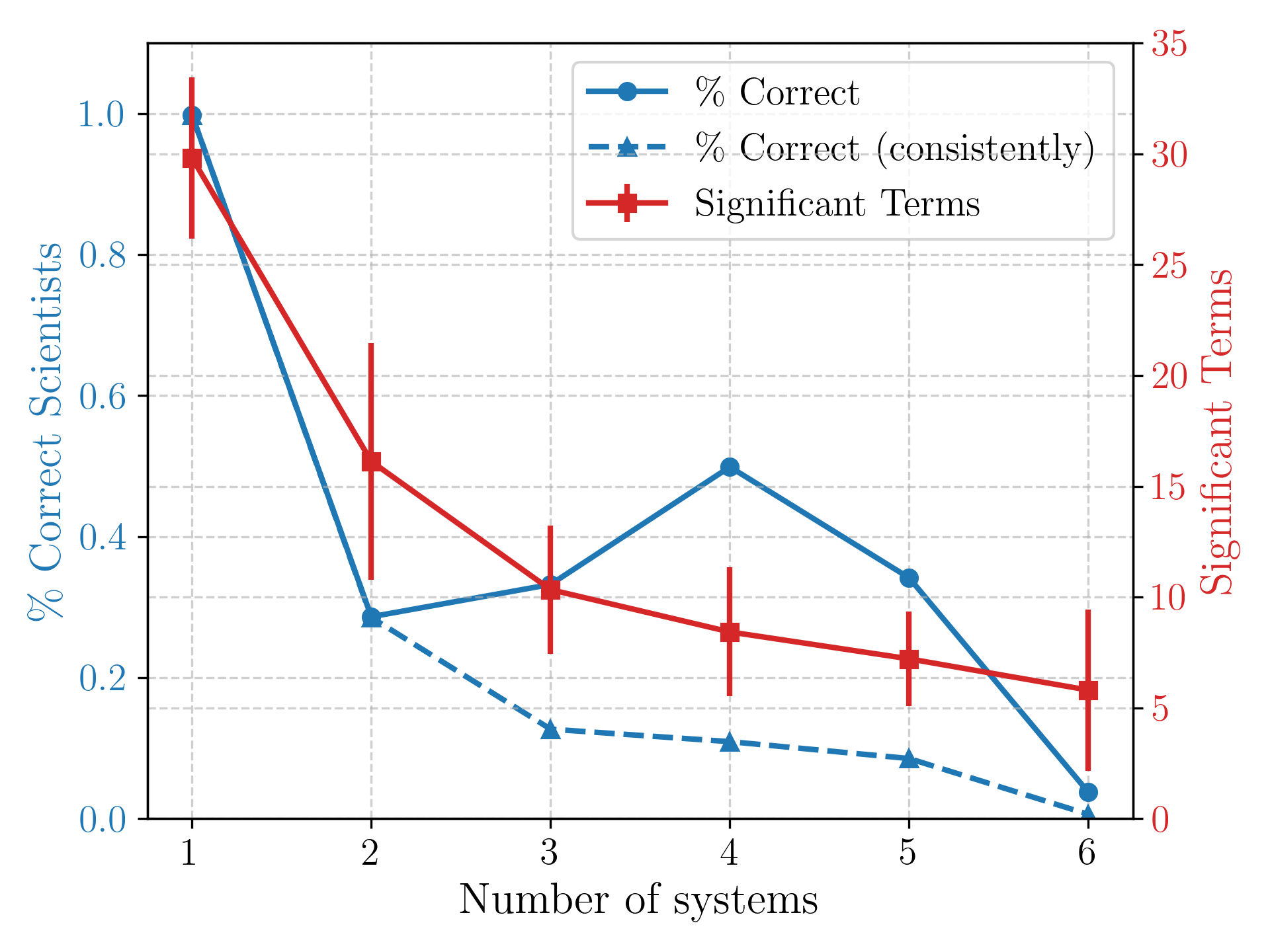}
    \caption{\label{fig:msms} Average number of significant terms and number of correct scientists as we increase the number of systems. Starting from the SHO, we include the pendulum, gravitational and relativistic harmonic problems at systems 2, 3, and 4 respectively, followed by two synthetic potentials (see Table~\ref{tab:systems}). The solid blue line (\% Correct) gives the percentage of seeds that such that the converged loss after the $n$-th training phase is less than $5 \times 10^{-3}$. The dashed blue line gives the percentage where the converged loss is less than $5 \times 10^{-3}$ for training phases up till the $n$-th one, i.e. MASS scientists that have always been correct. Results are parallelized over 1000 training seeds.}
\end{figure}

In the remaining of this section, we extend the analysis to a fully general case: multiple MASS scientists trained on multiple physical systems. Again, we train in the manner of Section~\ref{sec:method}, continuously exposing MASS to increasingly difficult systems and summing the errors across each system. 

Simultaneously, we present an extension of our MASS framework to unseen physical systems. Thus far we have been replicating the results of known problems: the simple harmonic oscillator, the simple pendulum, the gravitational potential and the relativistic oscillator. The original motivation for training MASS on these systems was that they were already well-studied, giving us a decent baseline to benchmark the performance of MASS against. However, a natural advancement in the direction of scientific progress, is what happens when we extend our current framework to systems yet to be discovered. At the same time, these four canonical systems lie far within the capabilities of MASS. The learnt theories are not very diverse (see Figure~\ref{fig:ms1s_corr}) and some terms in the final layer are almost consistently never used (see Figure~\ref{fig:strips}). Theoretically, this can be attributed to the fact that one-dimensional systems yield potential functions that typically do not involve the cross-terms $S_{xy}$. For example, even in the most complex relativistic harmonic oscillator that MASS has been exposed to can be expressed with a potential function following that of a Lagrangian 
\begin{align*}
    S = \mathcal{L} = \sqrt{1 - y^2} - \frac{1}{2} x^2
\end{align*}
for which $S_{xy} = 0$. 

To extend our studies to unseen physical systems and also fully utilize the capacity of the MASS network, we introduce synthetic systems. We list the modifications in Table \ref{tab:systems} by describing the kinetic energy $T$ and potential energy $V$ of each system. In particular, we introduce two additional synthetic systems which serve as extensions of the relativistic harmonic oscillator with a more complex potential energy term. 

\begin{table}[h!]
\centering
\caption{Summary of the seven 1D systems used in this work. For each system, 
we show the usual kinetic energy $T(x, y)$ and potential energy $V(x, y)$. The total energy is given by $T + V$. In this paper's convention, $\dot{x} = y$. The synthetic systems $\alpha, \beta$ are designed such that their first order Taylor expansions match the relativistic harmonic oscillator, up to addition and scaling by a constant. Note that in the relativistic cases (systems 4 to 6), the Lagrangian is not simply $T-V$ but the kinetic energy terms rather appears as $\gamma^{-1}$.}
\label{tab:systems}
\centering
\begin{tabular}{lcc}
\hline\hline
\textbf{System} & \boldmath$T(x,y)$ & \boldmath$V(x,y)$ \\
\hline
\textbf{(1) Classical} 
   & $\tfrac12\,y^{2}$ 
   & $\tfrac12\,x^{2}$ \\[6pt]
\textbf{(2) Pendulum} 
   & $\tfrac12\,y^{2}$ 
   & $1 - \cos(x)$ \\[6pt]
\textbf{(3) Kepler} 
   & $\tfrac12\,y^{2}$ 
   & $-\dfrac{1}{|x|}$ \\[6pt]
\textbf{(4) Relativistic} 
   & $\gamma$ 
   & $\tfrac12\,x^{2}\,\bigl(1 - y^{2}\bigr)^{3/2}$ \\[6pt]
\textbf{(5) $\alpha$ (Synthetic)} 
   & $\gamma$ 
   & $\tfrac12 x^2 \cdot \tfrac1\gamma$ \\[6pt]
\textbf{(6) $\beta$ (Synthetic)} 
   & $\gamma$ 
   & $\tfrac12 x^2 \cdot \cos(y)$ \\[6pt]
\hline\hline
\end{tabular}
\end{table}

Our key results are presented in Figure~\ref{fig:msms}, where we count the number of correct MASS scientists, defined as the number of seeds where the evaluation loss on the converged model, computed as the maximum MSE across all seen physical systems, is less than $5 \times 10^{-3}$. We also count the number of significant terms, defined as the number of terms in the final layer (out of $T=172$ terms) needed to reach 95\% of the total norm. These values are aggregated at the end of each training phase. Recall that in a single training phase, a MASS scientist is exposed to a new system and trained on the sum of the losses. Typically, a phase lasts for 10000 steps. 

As we increase the number of systems, the number of MASS scientists that have been consistently correct decreases (dashed blue line of Figure~\ref{fig:msms}), where to be consistently correct at phase $n$ is to have a low converged loss for all phases up till the $n$-th phase. This is intuitive, since the consistently correct MASS scientists at the end of the $n$-th phase is always a subset of that at the end of phase $n-1$. What is not very intuitive is the solid blue line: the number of correct scientists can increase with number of systems. This is analogous to seed 506 on Figure~\ref{fig:1sms_training}, where a MASS scientist can fail at a less complex system, but when exposed to more systems, learns the overarching underlying theory and succeeds. Such revivals of scientist networks highlight the importance of augmenting physical neural networks with more difficult tasks for it to work on simpler ones. 

The number of significant terms also show a consistently decreasing trend. This cements the results of Figure~\ref{fig:ms1s_corr} but is still surprising! To describe each system independently, the MASS scientist relies on rather different sets of weights, as in Figure~\ref{fig:strips}. Rather than learning separate terms to describe separate systems, i.e. learn the union of the terms for each theory, MASS instead learns the intersection of the terms, exemplifying the purpose of the shared final layer.

After training on 6 systems, the number of significant terms is still more than 6. A six term theory is neat, but nowhere near the simplicity of equations \ref{eqn:ham} and \ref{eqn:lag}. In the remaining of this section, we show that we can easily distill the underlying theory, and that this underlying theory is in fact a \textbf{Lagrangian}. 

\begin{figure}[ht!]
    \centering
    \subfigure[Fraction of each theory]{
        \includegraphics[width=0.22\textwidth]{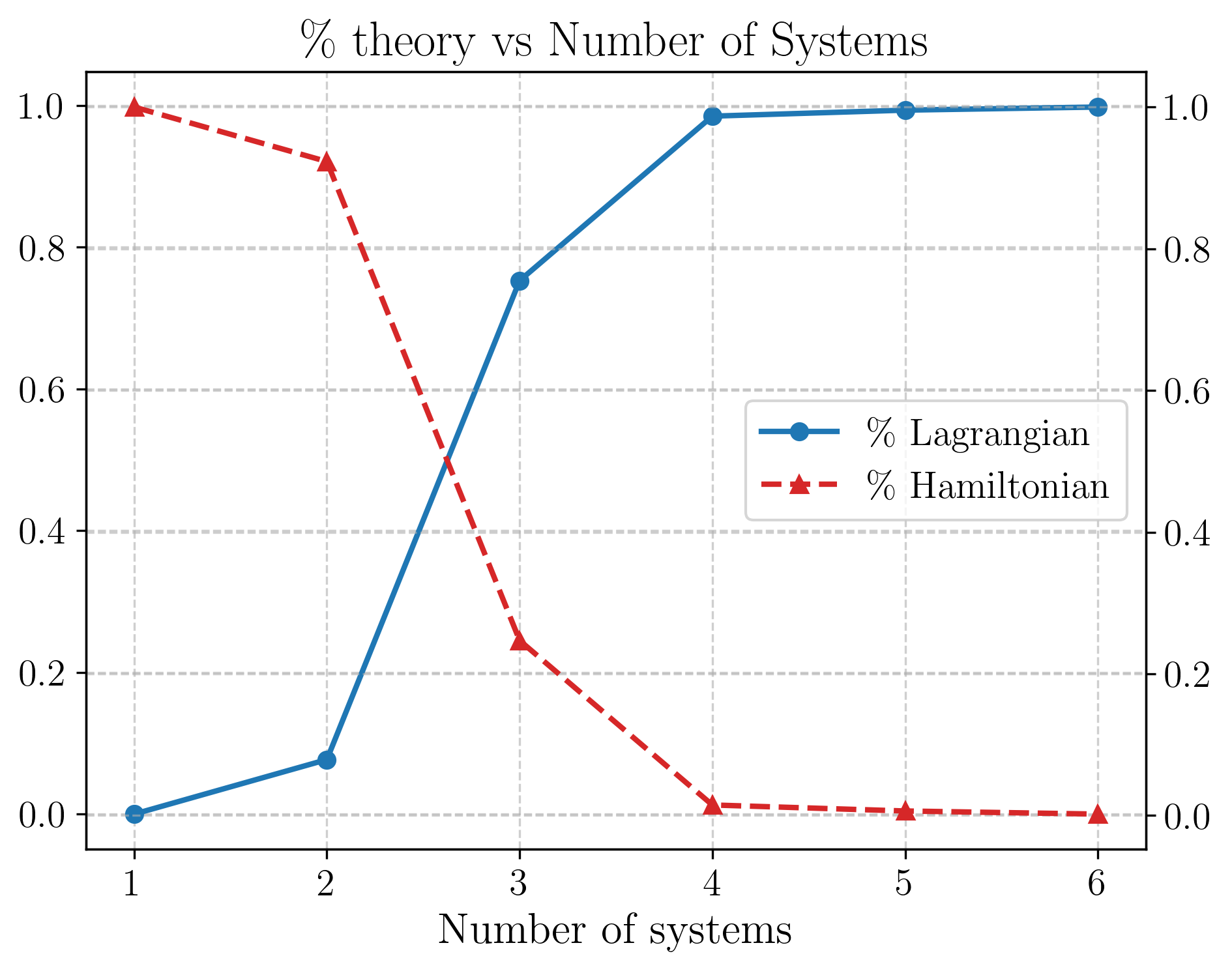}
    }
    \subfigure[$R^2$ of linear fitting]{
        \includegraphics[width=0.22\textwidth]{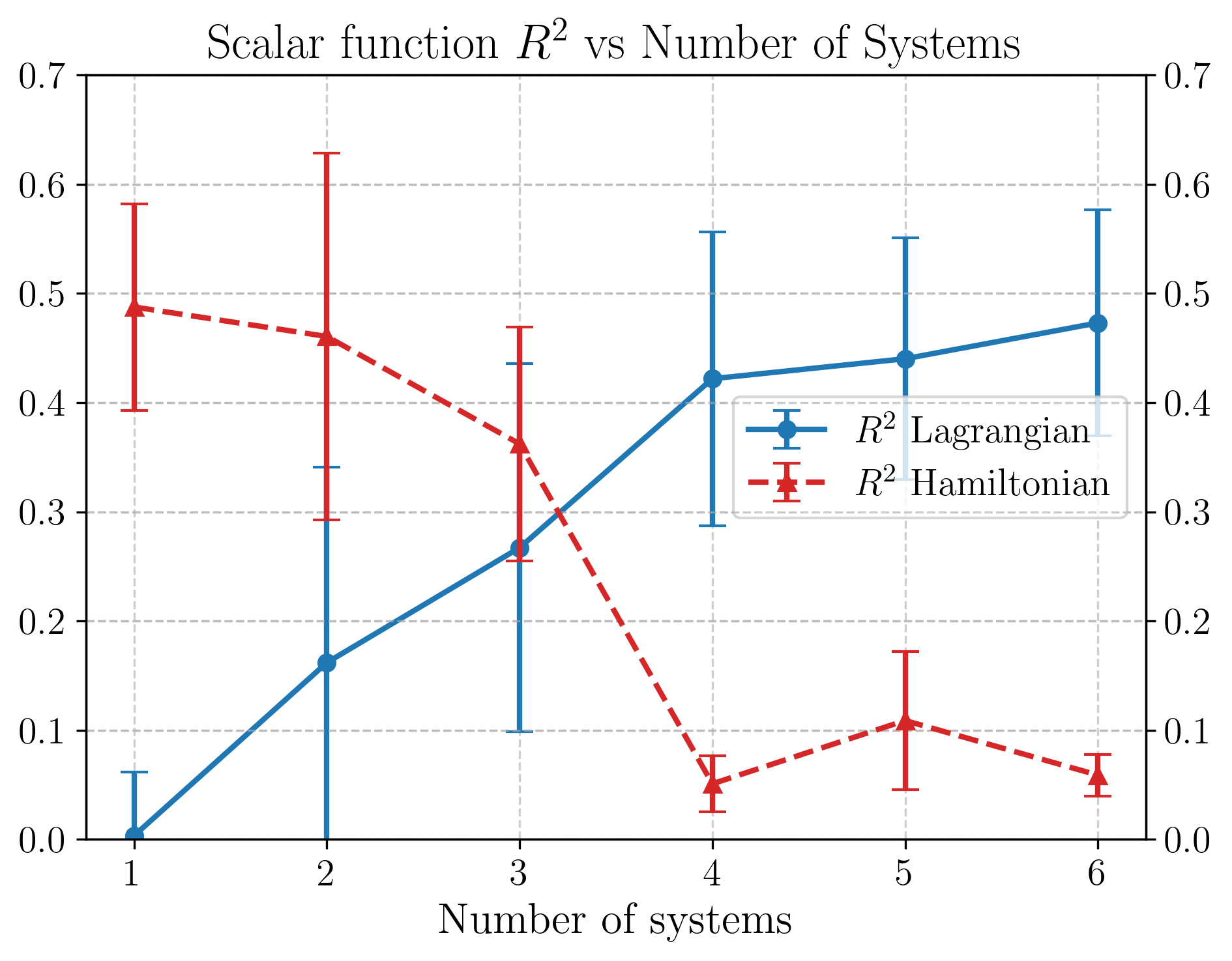}
    }
    \caption{MASS switches from learning a Hamiltonian theory to a Lagrangian one. \textbf{(a)} The fraction of MASS scientists that learns $c_1$ and $c_2$ to be opposite signs (Lagrangian) vs same sign (Hamiltonian). \textbf{(b)} $R^2$ score of linear fitting of activations to those derived by the Lagrangian vs Hamiltonian potential. The error bars show the standard deviation of the $R^2$ score.}
    \label{fig:theory_switch}
\end{figure}

\subsubsection{Simple Problems: Hamiltonian is all you need?}
Recall that in the Hamiltonian formulation, i.e. we learn $S$ to be $\mathcal{H} = T + V$, where $\mathcal{H}$ is the Hamiltonian, $T$ is the kinetic energy, $V$ is the potential energy. In the Lagrangian formulation, we learn $S$ to be $\mathcal{L} = T - V$. The sign flip here is crucial. 

Given data coordinates $x_i, y_i$, and weights $\theta$ of the MASS scientist, we can compute the scalar function $S(x_i, y_i)$. We can also pre-compute the kinetic and potential energy terms $T_i$ and $V_i$, then linear fit $S$ with $c_1 T + c_2 V$. We denote a MASS scientist to have learnt a Lagrangian theory if $c_1$ and $c_2$ have opposite sign and a Hamiltonian theory if $c_1$ and $c_2$ have the same sign. 

Alongside the discrete counting method described above, we can also directly fit a batch of activations against the Lagrangian and Hamiltonian activations which we can compute from an analytic expression. This fitting should not be expected to be perfect, for MASS can learn a simple variant of a clean theory yet give similar accuracies. For example, learning $S$ and $S + x$ may end up being effectively the same since terms second first derivative terms change by a constant while second derivative terms are entirely the same. Despite the imperfections of linear fitting, the aggregate trend of the mean $R^2$ across many samples can tell us a bit about the relation to each theory.


Figure~\ref{fig:theory_switch} summarizes these results and shows the evolution of theories learned across a number of systems. When trained on just the simple harmonic oscillator or the pendulum, MASS learns almost a complete Hamiltonian description (with more than 90\% of the scientists agreeing). In this simple setting, there exists some choice of sparse terms among the $T=172$ derivative terms that under Hamilton's equations (Equation \ref{eqn:ham}) that give low loss, and MASS tends towards this. The learned scalar functions themselves also display strong correlation. 
\subsubsection{Complex Problems: Lagrangian is all you need}
The story changes when we extend beyond the simple pendulum to more complex problems. On these systems (3 to 6 in Table \ref{tab:systems}), MASS switches to a Lagrangian theory. One reason for this, as discussed in \cite{lnn} is that the Lagrangian can be applied directly in generalized coordinates while the Hamiltonian requires canonical coordinates. As our data is presented in generalized coordinates, the MASS architecture supports calculations done in this coordinate system, following that of the Lagrangian formulation.  What is surprising here is that the correlation to the Lagrangian scalar function itself also increases, suggesting that on an aggregate level, AI scientists tend towards this singular family of descriptions of physical systems: the Lagrangian description!

The results of Figure~\ref{fig:theory_switch} show a bias toward the Lagrangian formulation, but never a definitive proof that the calculations faithfully follow that of the Lagrangian. Of course we should not expect that, given the capacity imbued to MASS, why would it follow some ``nice" theory? \textbf{But turns out, it almost exactly does!} We will show this with a method of constrained optimization. 

In the Lagrangian formulation, the prediction of $\dot{y}$ will be given by \citep{lnn} 
\begin{align*}
    \dot{y} = S_{yy}^{-1} (S_x - S_{xy} y).
\end{align*}
The activations in the final layer of MASS will hence be concentrated on the terms $S_{yy}^{-1} S_x$ and $S_{yy}^{-1} S_{xy} y$. However, the multi-expressivity of our network allows for many terms to be linearly related to these two terms. 

\begin{table}[h!]
\centering
\caption{Constraint optimization on the objective of Equation~\ref{eqn:optim}. 
The goal is to reduce the activations for predicting $\dot{y}$ to either one or two terms. 
High $R^2$ values for the Lagrangian indicate that the learned network recovers the same 
functional dependence as the analytical Lagrangian, just embedded in higher dimensions.}

\label{tab:constraint}
\begin{tabular}{lcccc}
\hline\hline
\textbf{System} 
    & \boldmath{$R^2_\mathcal{L}$} 
    & \boldmath{$R^2_\mathcal{H}$}\\
\hline
\textbf{(1D) Relativistic} 
    & \textbf{0.9999}
    & 0.9995 \\[6pt]
\textbf{(1D) $\alpha$ (Synthetic)} 
    & \textbf{0.9835}
    & 0.8205 \\[6pt]
\textbf{(1D) $\beta$ (Synthetic)} 
    & \textbf{0.9306}
    & 0.8734 \\[6pt]
\textbf{(2D)} \textbf{Double pendulum} & \textbf{0.9712} & 0.7317 \\
\hline\hline
\end{tabular}
\end{table}

We solve the constrained optimization problem. Given data coordinates $x_i, y_i$, and weights $\theta$ of the MASS scientist $\mathcal{M}$, we can compute the scalar function $S(x_i, y_i)$ and from this obtain two terms representative of the Lagrangian theories. We call these $u_i = S_{yy}^{-1} S_x$, and $v_i = S_{yy}^{-1} S_{xy} y$. $u_i$ and $v_i$ can be easily computed with JVP. We can also obtain the activations $\textbf{a}_i$ of the final layers with a forward pass of $(x_i, y_i)$ through $\mathcal{M}_\theta$. Then, we solve the constrained optimization problem
\begin{align} \label{eqn:optim}
    \min \; & \mathbb{E}[(\hat{u}_i - u_i)^2 + (\hat{v}_i - v_i)^2] \\
    s.t. \; & \hat{u}_i = \sum_{j = 1}^{T} c_{j} a_{ij} \\
    & \hat{v}_i = \sum_{j = 1}^{T} d_{j} a_{ij} \\
    & c_j + d_j = 1 \; \forall j \in [1, T] \label{constraint}
\end{align}
where $\hat{u}_i, \hat{v}_i$ is a transformation of the 172 term MASS activation space to the 2 term Lagrangian activation space, and the constraint \ref{constraint} restricts the transformation to one which exactly uses all the weights in the activations $\textbf{a}_i$, i.e. no cheating by placing avoiding some activations completely and overusing others. Just a technical note: in the first four systems of Table \ref{tab:systems}, we always get a trivial solution of $\hat{v_i} = 0, c_j=1$ since $v_i = 0$ (due to the cross term $S_{xy} = 0$. Of interest is what happens in the synthetic systems, where the cross terms are not zero and MASS is forced to learn something non-trivial in both $u$ and $v$. 

We summarize these results in Table~\ref{tab:constraint}, which consists of the single term $-S_x$. We average the $R^2$ scores of this constrained optimization fitting across the correct scientists to give Table~\ref{tab:constraint}. Coherent with previous observations, MASS can almost be directly transformed into a Lagrangian theory with $R^2$ values above 0.9. If we try to pick any two random terms from the $T=172$ available terms, or even the two terms with the highest activation magnitudes, the constrained optimization will typically fail, observed as a negative $R^2$ score on the holdout test set. 

Such strong correlations to the Lagrangian raises a broader question: can we find a third description of classical mechanics? At least with MASS working in the rich theory space of $T=172$ terms, the answer appears to be no! The Lagrangian is all you need.

\subsection{Extensions to high dimensions}

\begin{figure}[h!]
    \centering
    \includegraphics[width=\linewidth]{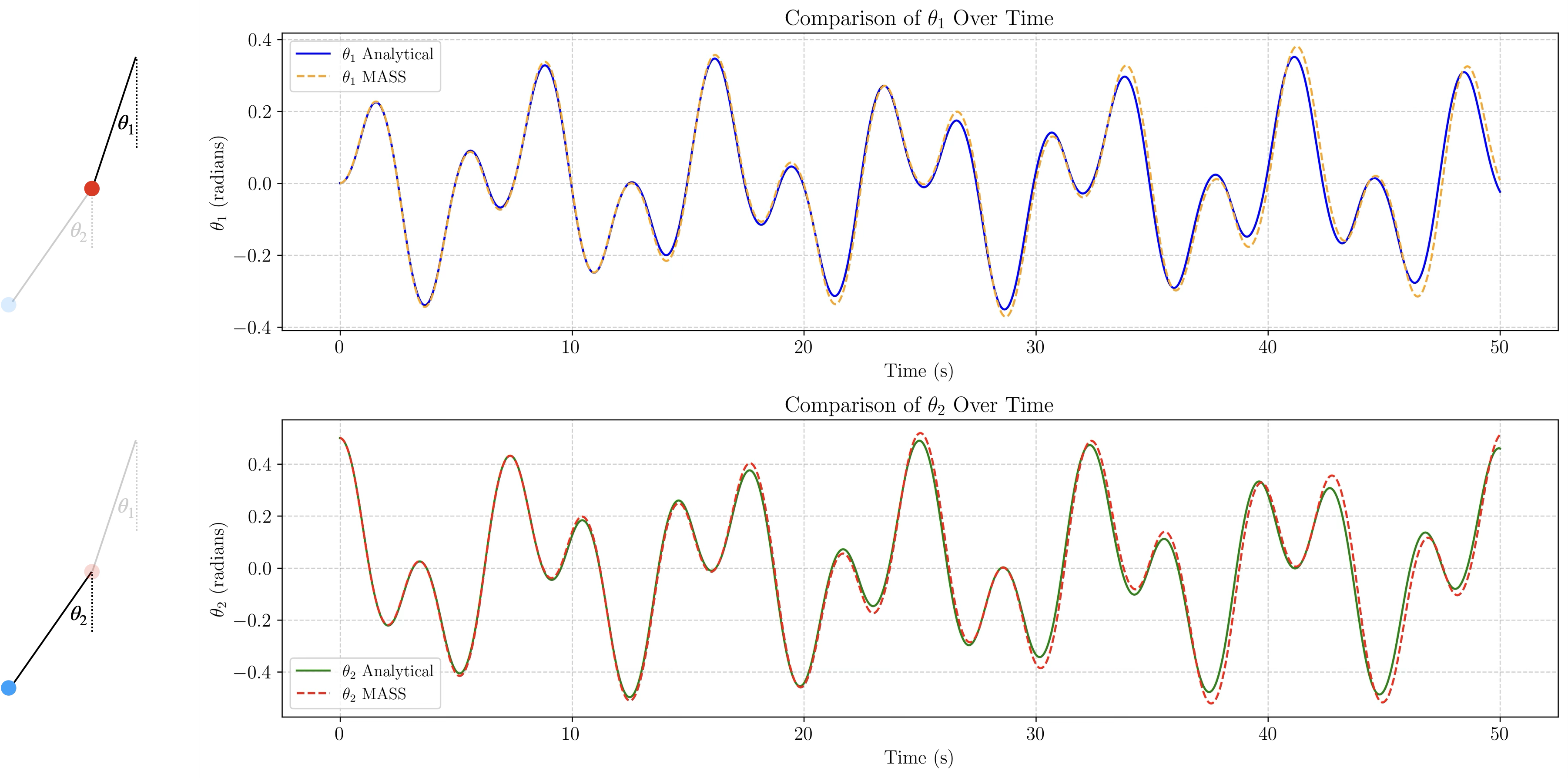}
    \caption{\label{fig:double_pendulum} Trajectories of the double pendulum solved by MASS to an MSE of $5 \times 10^{-3}$. ODE is solved with RK4 integration with a time-step of $dt=0.05$.}
\end{figure}

While in the previous sections, we have mainly worked with one-dimensional problems, i.e. $x \in \mathbb{R}$, most physical problems in nature are higher dimensional. In this section, we study one classic example of that: the chaotic double pendulum problem. The two degrees of freedom are the angles of the two pendulums. Our results show that MASS can be effectively extended to higher dimensions.

Following an identical training scheme as in Section~\ref{sec:method}, we reproduce the analytically correct trajectory of a double pendulum in Figure~\ref{fig:double_pendulum}, calling the MASS solver for $\dot{\mathbf{y}}$ at each step and using RK4 integration~\citep{runge1895, kutta1901}. 

Not only can we achieve rather accurate prediction of the angles, the energy discrepancy is only at 0.4\% of the total energy per 100 steps. This is comparable to the results of the Lagrangian neural network~\citep{lnn}. Even without introducing the Lagrangian and Euler-Lagrange equations directly into the architecture to enforce energy conservation, MASS learns to reproduce it.

We also observe, consistent with our expectations, that the learnt theories resemble a Lagrangian, with the results further included in Table~\ref{tab:constraint}. 

We present more results of solution trajectories to the spherical pendulum and the multi-body gravitational problem in Appendix~\ref{sec:high_dim}. We are not claiming that MASS as a state-of-the-art method for solving physical systems, especially since it is out of the scope of this project to tune MASS for efficiency and accuracy on higher dimensional problems. In fact, the computation of the Hessian matrix and its inverse incurs an $O(d^2)$ dependence on the dimensionality $d$ of the problem, so directly applying the current solver to problems of extremely high dimensionality would be very expensive. Nonetheless, the evident applicability of MASS to solving the double pendulum problem at a sufficient level demonstrates its potential for future exploration, and drives home the spirit of this paper: to build AI scientists that are both \textbf{simple and interpretable}, and also \textbf{generally applicable} to complex physical systems.

\section{Discussions}
So do two AI scientists agree? The short answer is \textbf{yes}. But it comes with some caveats. 

Looking back, we question the relationship between our results in Figure~\ref{fig:ms1s_corr} and those in Figure~\ref{fig:theory_switch}. In the former, we observe a strong correlation between the theories described by each MASS scientist. Compared to Figure~\ref{fig:theory_switch}, we see an indication that scientists can learn different theories. In combination, this is suggestive that a number of theories lie on the boundary between a ``Hamiltonian”-like contour and a ``Lagrangian”-like contour. We did not perform rigorous symbolic regression on the results of the learned scalar function. Given the vast number of terms that can be learned by MASS, we believe the results of Figure~\ref{fig:theory_switch} and Table~\ref{tab:constraint} tell a much richer story about the underlying theories. We performed thorough numerical analysis of the trained systems through counting the number of Hamiltonian and Lagrangian theories, and measuring correlations of the activations, to conclude the generalizability of the Lagrangian theory. 

To answer the original research question, we chose to use different seeds as proxies for different AI scientists. While only affecting the initializations of the MASS network, we already see strikingly different training behavior (Figure~\ref{fig:1sms_training}). Our initial experiments on varying model width and depth suggest that the extent of agreement increases with model capacity. Preliminary testing on varying architectures, using convolutions and attention instead of pure feed-forwards, shows to be much less stable in training, primarily due to the low-dimensionality of our data.

Looking ahead, while most of our results, due to the extensive parallelization over many seeds and systems, were conducted on one-dimensional physical systems, preliminary results (Figure~\ref{fig:double_pendulum}) show that these can be readily extended to higher dimensions. 

MASS offers a tradeoff between inductive bias (through including physical priors such as the Euler-Lagrange equations) with training efficiency. When calculating many of the $T=172$ terms, particularly the Hessian matrix inverses, training slows and becomes unstable, which was solved only with strong regularization and initialization techniques. Nonetheless, these additional terms should not be seen as irrelevant. One should not expect the Euler-Lagrange equations to be the end-all for physics-based machine learning, and certainly not for physics itself.

\section{Future Work}
There are several low-hanging fruits to extend our work in this paper. We list several below: 
\begin{enumerate}
    \item \textit{Coordinate choice.} Our experiments done in generalized coordinates forbid the Hamiltonian expression to achieve low loss, while the Lagrangian remains a perfect description. Hence, results such as Figure~\ref{fig:theory_switch} are not extremely surprising. But what happens if we allow MASS to work in arbitrary coordinates? This can be done by allowing MASS to learn a coordinate transform (through a simple MLP) then take derivative in the transformed coordinates \citep{hnn_transform}. On these coordinates, will MASS still prefer the Lagrangian expression?
    \item \textit{Loss function.} We can modify the loss function to encourage the learning our un-learning of specific theories. In particular, the measure of Hamiltonicity~\citep{hidden_sym} quantifies how "Hamiltonian"-like the theory is. How does including this as a loss term bias MASS into learning different theories? 
    \item \textit{Model architecture.} Our choice of variation of AI scientists was across the random initializations. What happens if we modify the architecture completely? Will AI scientists still agree? 
    \item \textit{High dimensions.} We show results up to six dimensions in Figure~\ref{fig:three_body}. But many physical problems are even higher dimensional. How do we efficiently extend our model to solve those problems?
\end{enumerate}

\section{Conclusion}
In this paper, we have developed a novel architecture and training scheme, \textbf{MASS} and rigorously investigated the evolution of theories studied by MASS across multiple physical systems. Through our experiments, we show that AI scientists, when modeled as a high capacity neural network, often learns multiple equivalent expressions of the same theory. As we expose our AI scientists to new, and more complex systems, some of these theories prove inconsistent with these previously unseen systems, while others successfully generalize to more difficult problems. Even within these surviving theories, the underlying theories change over increasing systems, starting from resembling a Hamiltonian to resembling a Lagrangian.

We hope that MASS will not just be an interesting story of Hamiltonian v.s. Lagrangian, but also lays the groundwork to build models that are more interpretable and capable. Then, we will revisit the question: \textbf{do two AI scientists agree?}

{\bf Acknowledgement} Z.L. and M.T. are supported by IAIFI through NSF grant PHY-2019786. Z.L. is supported by the Google PhD Fellowship.

\bibliography{main}

\appendix

\section{Training methods} \label{appendix:hyper}
\begin{table}[h!]
\centering
\caption{Hyperparameter settings for training MASS.}
\label{tab:hyper}
\centering
\begin{tabular}{lcc}
\hline\hline
\textbf{Parameter} & Value \\
\hline
    MLP hidden layers & 4\\
    MLP width & 20\\
    Batch size & 512 \\
    Steps (per phase) & 10000\\
    Linear warmup steps & 100 \\
    Learning rate & $5 \times 10^{-4}$ \\
    Weight decay & 0.01 \\
    $\beta_1$ & 0.7 \\
    $\beta_2$ & 0.8 \\
    EMA & 0.99 \\
    $\lambda_b$ & 0.5 \\
    $\lambda_1$ & 0.1 \\
    $\lambda_2$ & 0.01 \\
\hline\hline
\end{tabular}
\end{table}

As stated in Section~\ref{sec:method}, training MASS is extremely unstable. The ``ground truth" scalar potentials in many of the systems(Table~\ref{tab:systems}) we are studying induce Hessian matrices that are identically zero, making their inverses hard to compute. To remedy this, we introduce a regularization parameter $b_j$ to each $S_j$, computing $\rm{pinv}(\mathbf{A} + b)$ instead of $\rm{inv}(A)$, and minimize the loss with the inclusion of a $\lambda_b ||b||_1$ term to each system. We consistently use $\lambda_b = 0.5$ in our experiments. In addition, we augment the initializations. Under a typical initialization scheme, like Kaiming initialization~\citep{kaiming_init} and Xavier initialization~\citep{xavier_init}, the second derivatives are very small, leading to the same problem of exploding inverses. Instead of symbolically optimizing for the variances at each layer~\citep{lnn}, we simply augment the input of the MLP to take not just $(\mathbf{x}, \mathbf{y})$ but the entirety of $(\mathbf{x}, \mathbf{y}, \mathbf{\dot{x}}, \mathbf{\dot{y}}, \mathbf{xy})$. Together, these allow for stable training of MASS networks up to high learning rates of even $1 \times 10^{-2}$. 

To encourage sparsification of terms, we introduce a regularization term on the final layer weights and activations. Note that regularizing the weights alone is not enough, since MASS can simply cheat by increasing the magnitude of $S$ and increasing the activations. Let the final layer weights be $\mathbf{w}$ and the activations be $\mathbf{a}_j$ for system $j$, then system-wise we include the regularization term $\frac{\lambda_1}{T} ||\mathbf{w}||_1 + \frac{\lambda_2}{T} ||\mathbf{a}_j||_1$. We use $\lambda_1 = 0.1, \lambda_2=0.01$ in our experiments.

We report the hyperparameter settings in the Table~\ref{tab:hyper}. 

For the higher dimensional problems, we used larger MLP width ranging from 40 to 100 and up to 6 hidden layers. Correspondingly, the learning rate varied between $5 \times 10^{-4}$ to $5 \times 10^{-3}$.

\section{Activations on different 1D systems}\label{sec:1d_acts}
\begin{figure}[h!]
    \centering
    \includegraphics[width=\linewidth]{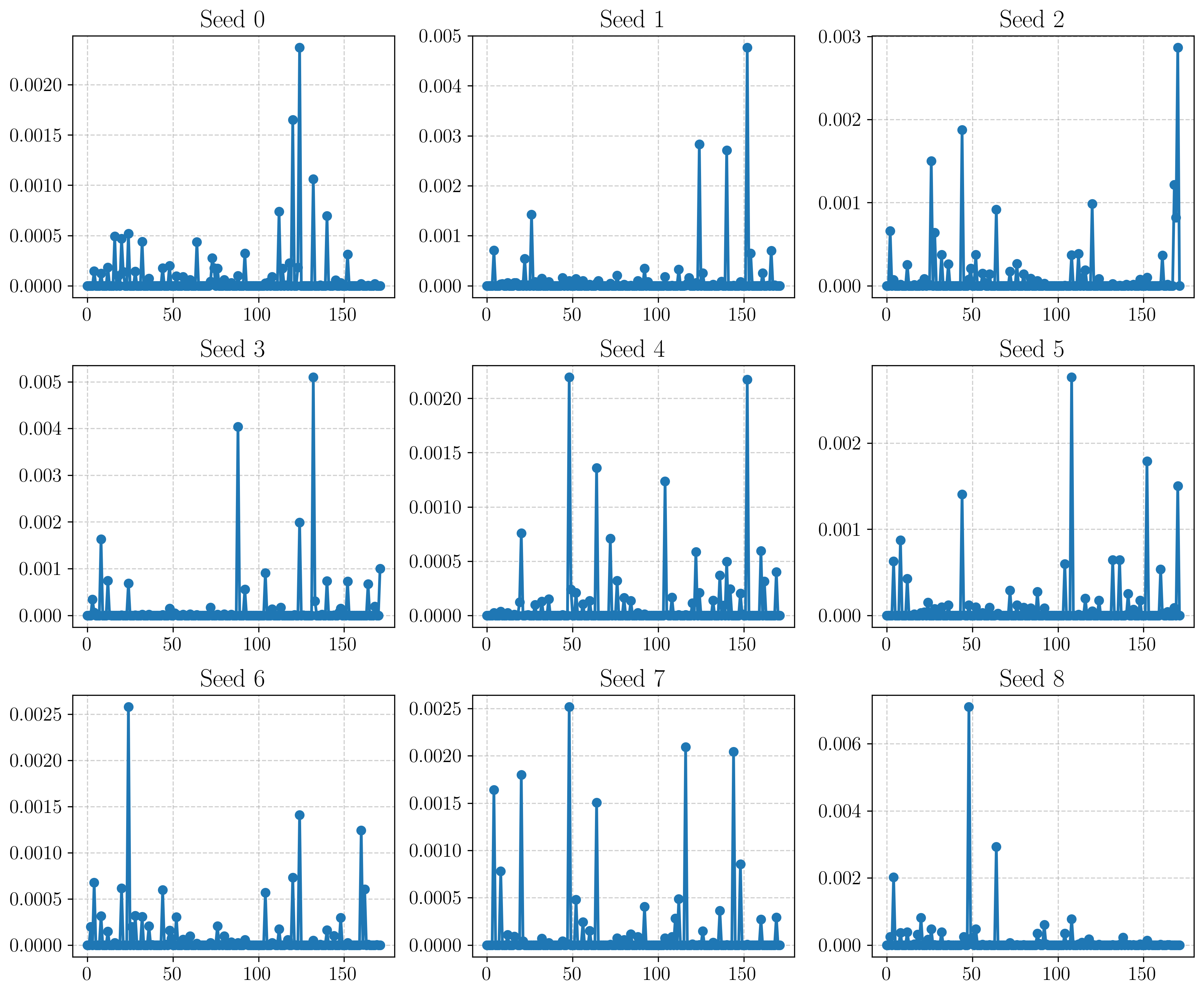}
    \caption{\label{fig:multi_scientists_classical_acts} Mean absolute activations of multiple scientists when trained on the same system: simple harmonic oscillator. Exact activation magnitudes differ, and many terms are activated in general.}
\end{figure}

\begin{figure}[h!]
    \centering
    \includegraphics[width=0.8\linewidth]{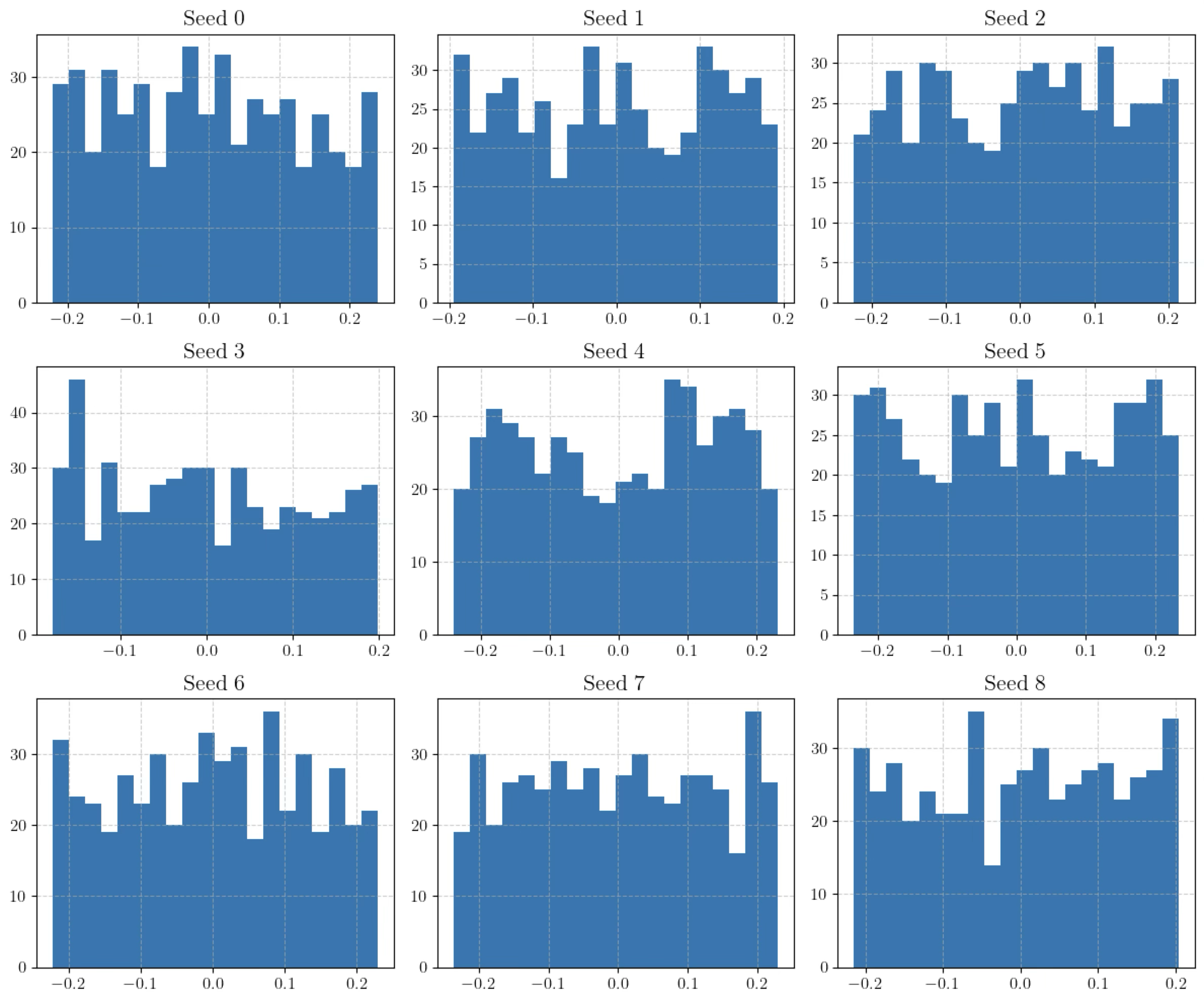}
    \caption{\label{fig:multi_scientists_classical_hist} Mean absolute activations of multiple scientists when trained on the same system: spring-mass.}
\end{figure}

\begin{figure}[h!]
    \subfigure[Relativistic]{
    \label{fig:multi_scientists_relativistic_hist}
        \includegraphics[width=0.4\linewidth]{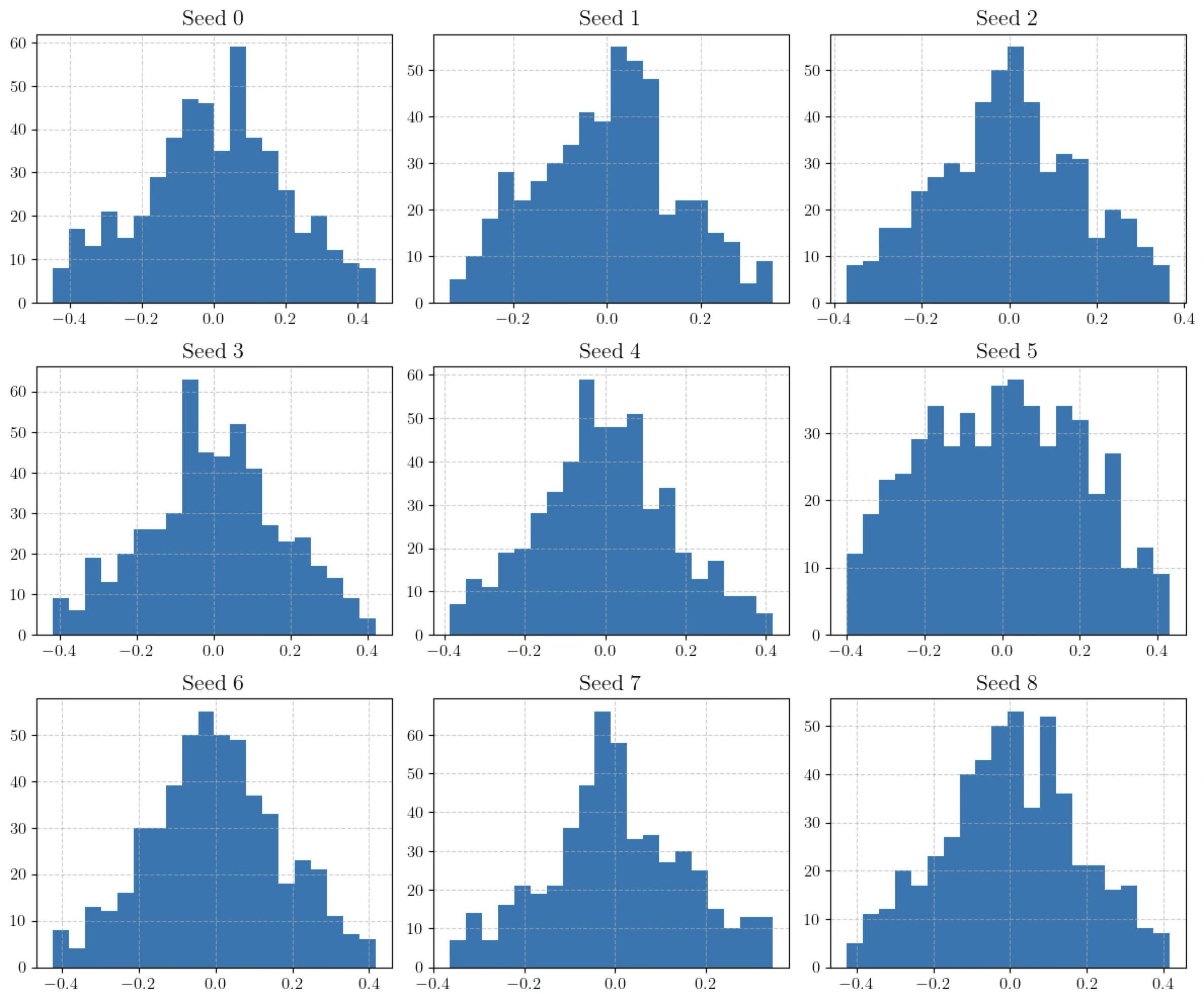}
    }
    \subfigure[Simple pendulum]{
    \label{fig:multi_scientists_pendulum_hist}
        \includegraphics[width=0.4\linewidth]{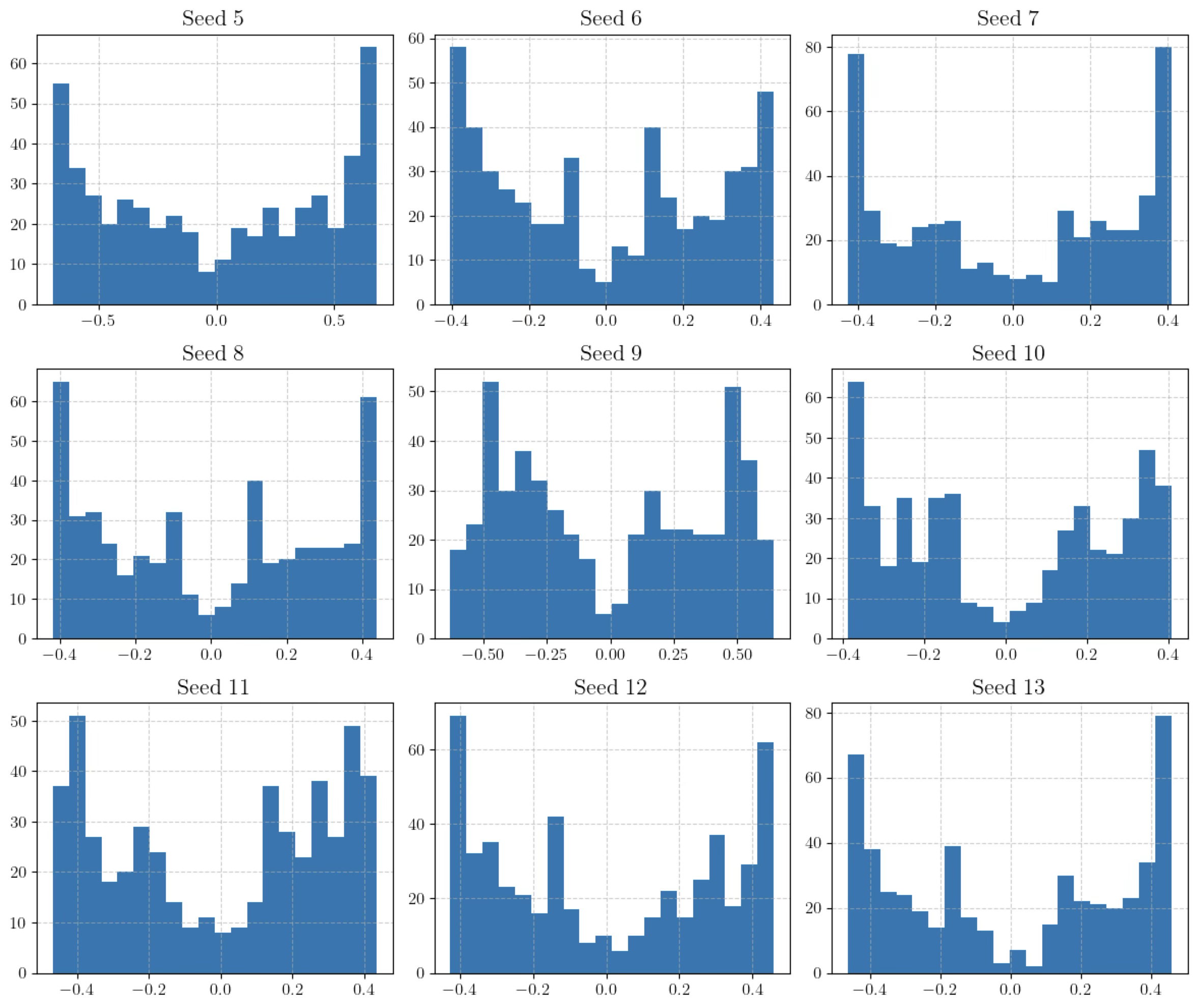}
    }
    \caption{Mean absolute activations for multiple MASS scientists when trained on the same single system, either \textbf{(a)}the relativistic harmonic oscillator or $\textbf{(b)}$ the simple pendulum}
    \label{fig:multi_scientists_hist}
\end{figure}

In the following set of visualizations (Figures~\ref{fig:multi_scientists_classical_acts} to \ref{fig:multi_scientists_pendulum_hist}), we support our claims that while the exact terms learnt by MASS differ, the underlying theories, described by the histograms, are mostly identical. We find that this first PCA component corresponds mainly to the ground truth acceleration for 1D simple systems (1 to 4), but not necessarily for more complex systems. It is of future interest to investigate that this direction means and what the theories that have a low correlation to this direction actually represent.

\newpage

\section{Visualizing learned scalar functions}\label{appendix:scalar}

\begin{figure}[h!]
    \centering
    \subfigure[1.1]{
        \includegraphics[width=0.12\textwidth]{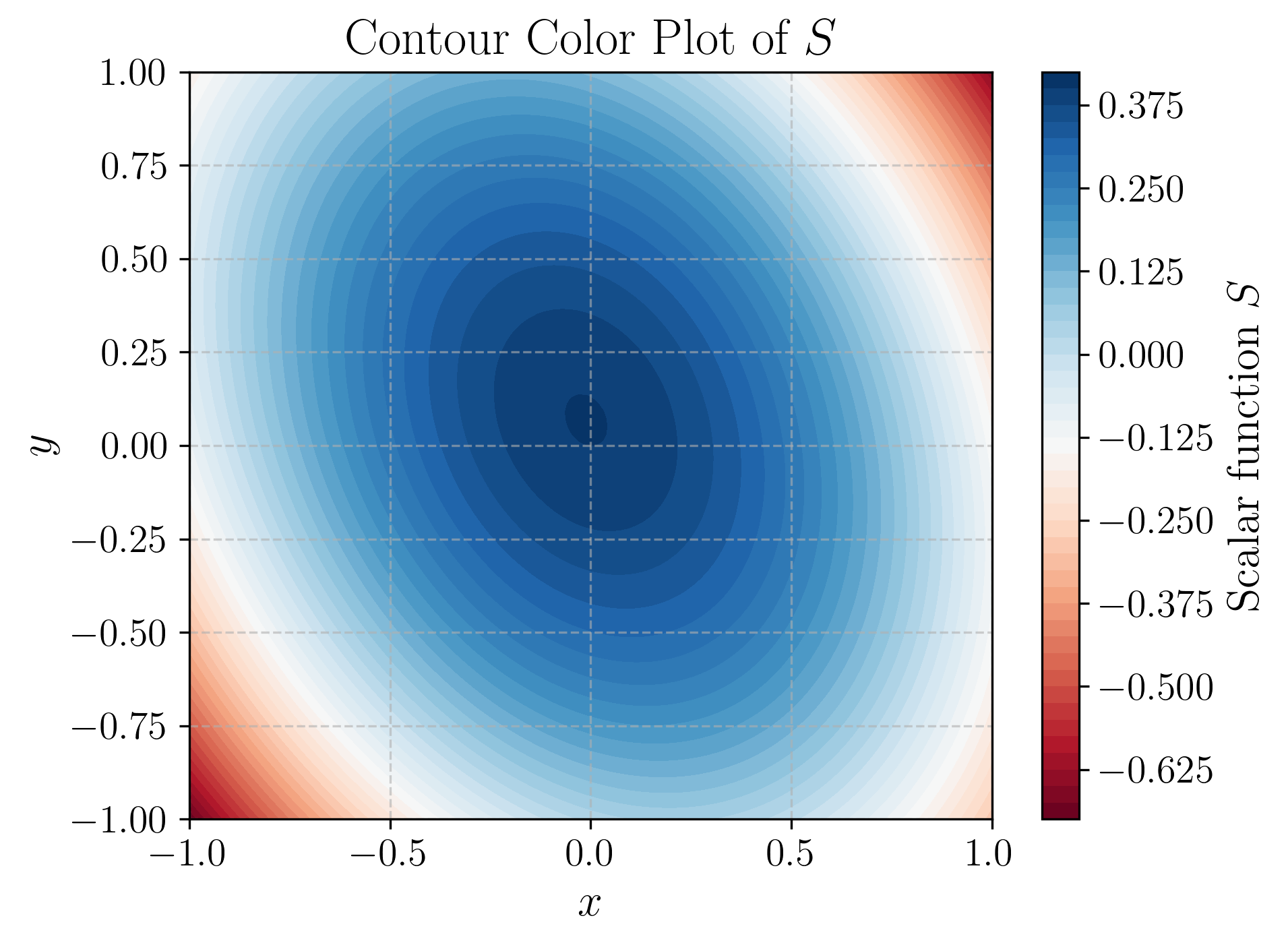}
    }
    \subfigure[1.2]{
        \includegraphics[width=0.12\textwidth]{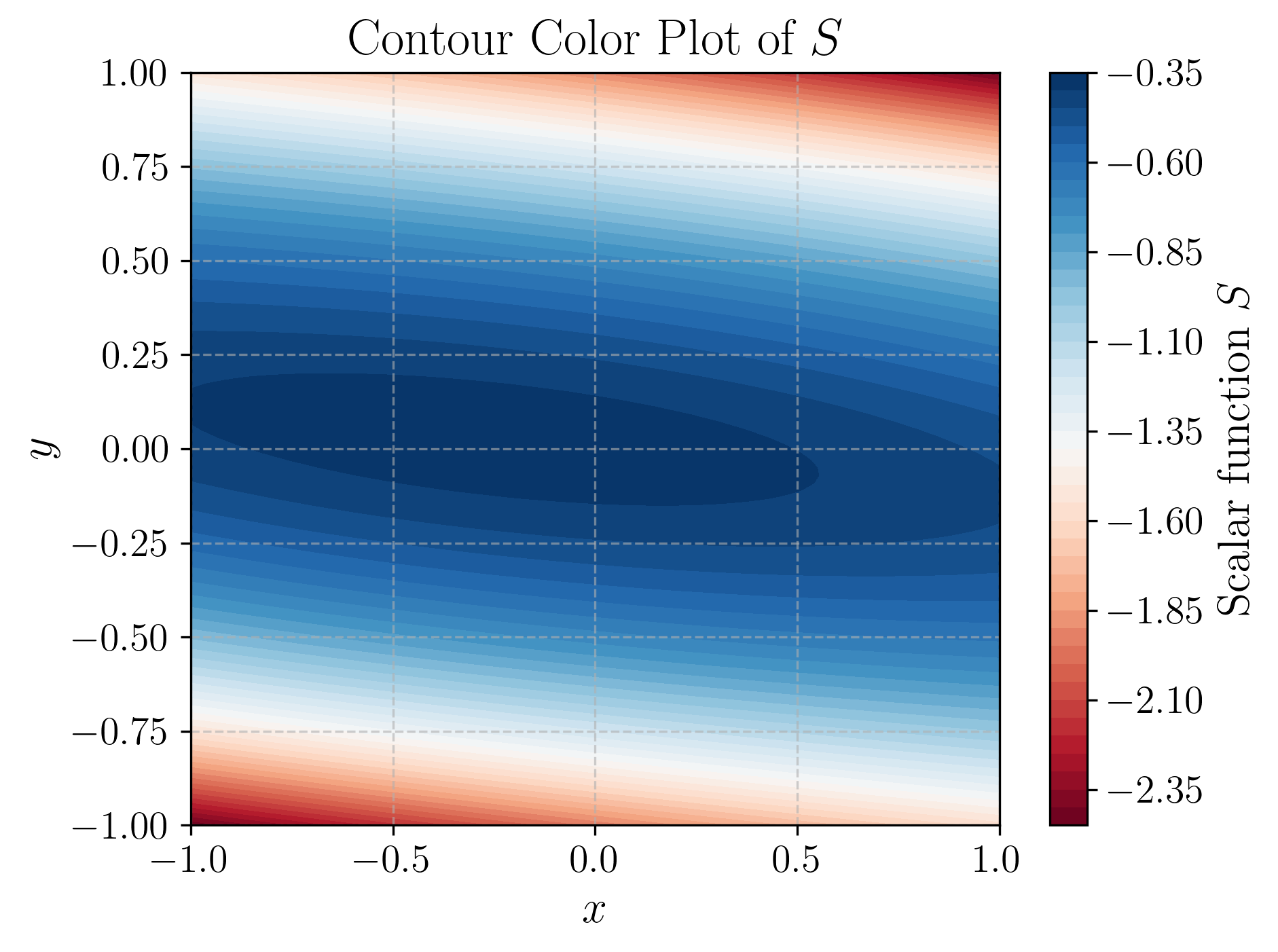}
    }
    \subfigure[1.3]{
        \includegraphics[width=0.12\textwidth]{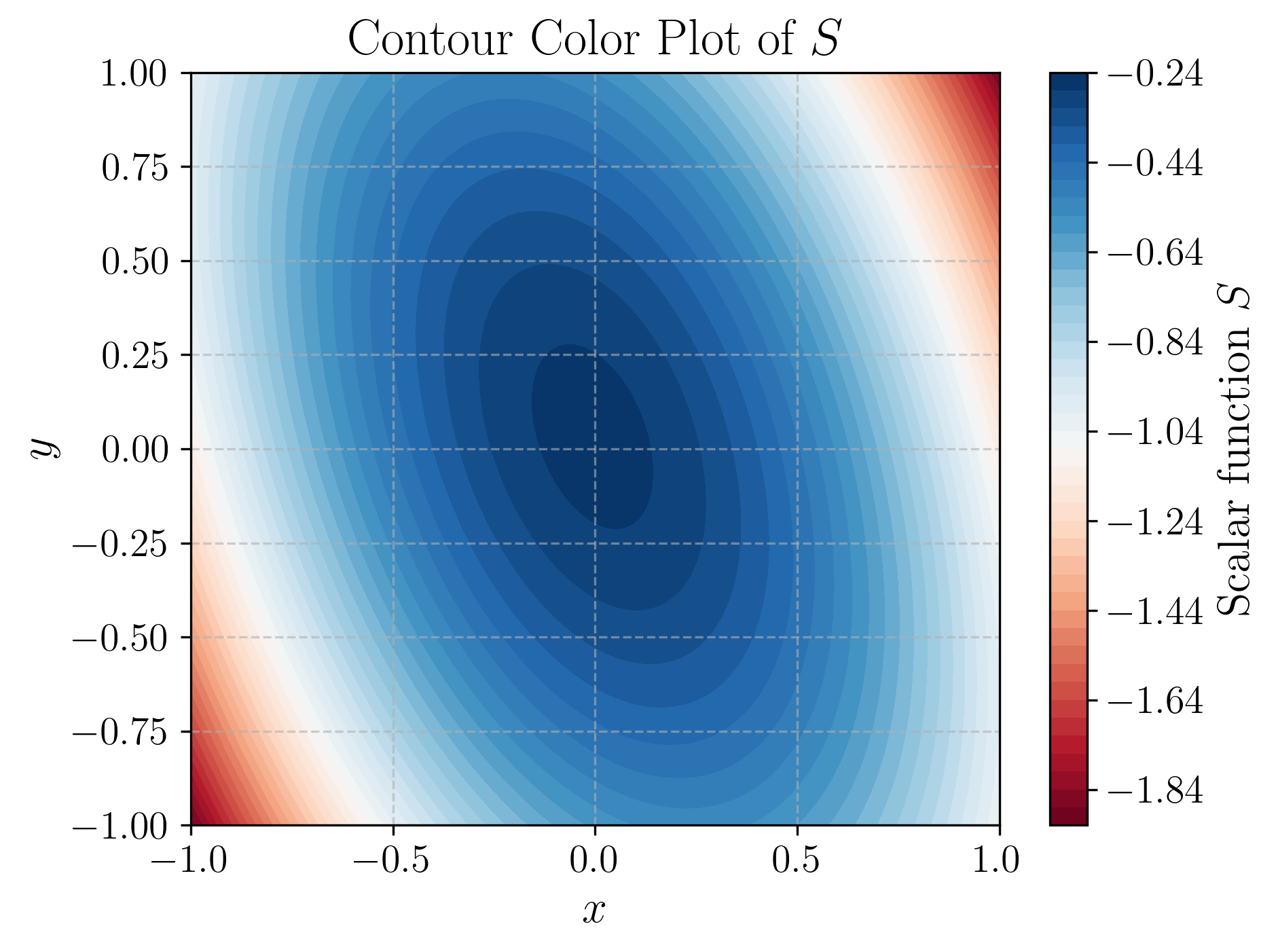}
    }
    \subfigure[2.1]{
        \includegraphics[width=0.12\textwidth]{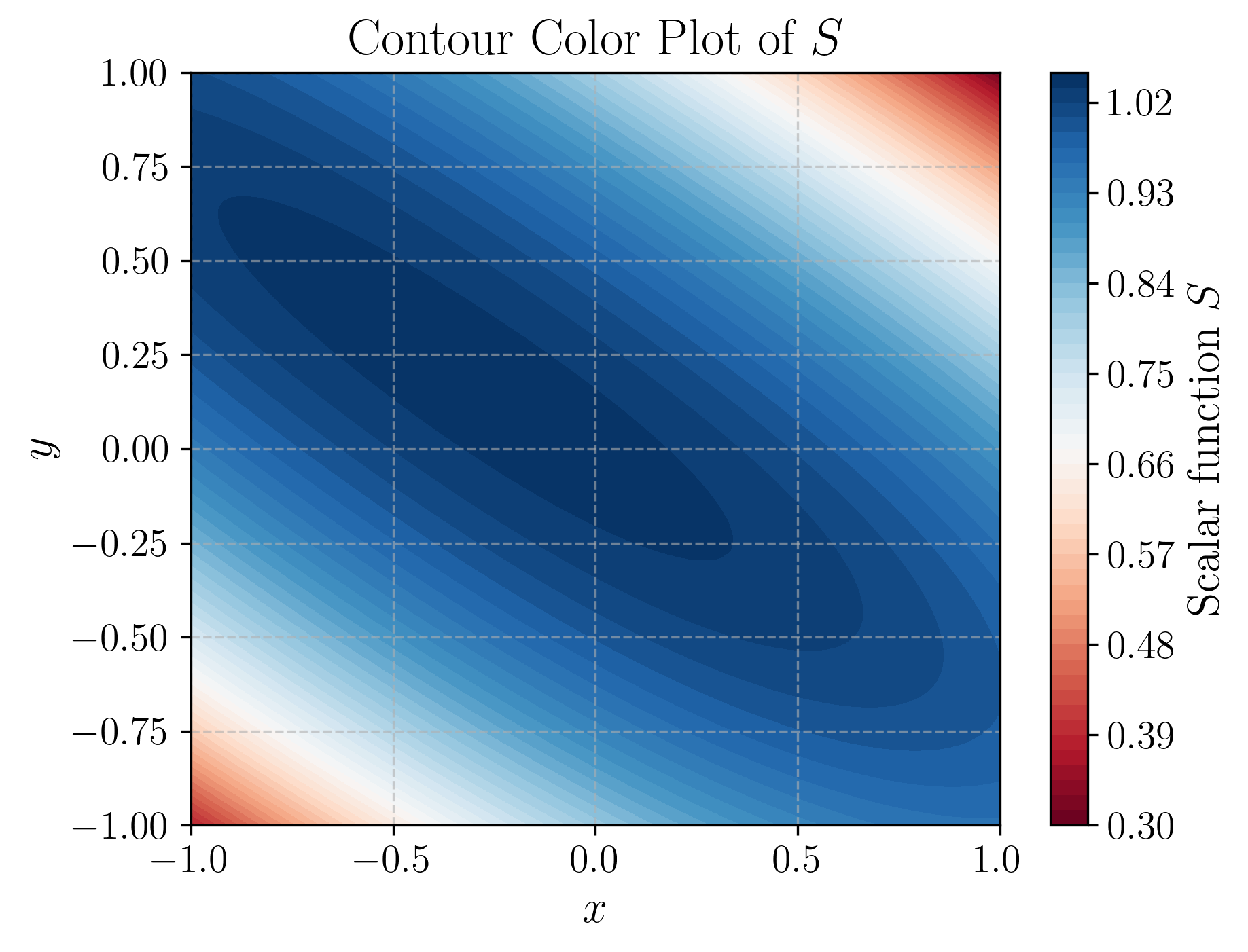}
    }
    \subfigure[2.2]{
        \includegraphics[width=0.12\textwidth]{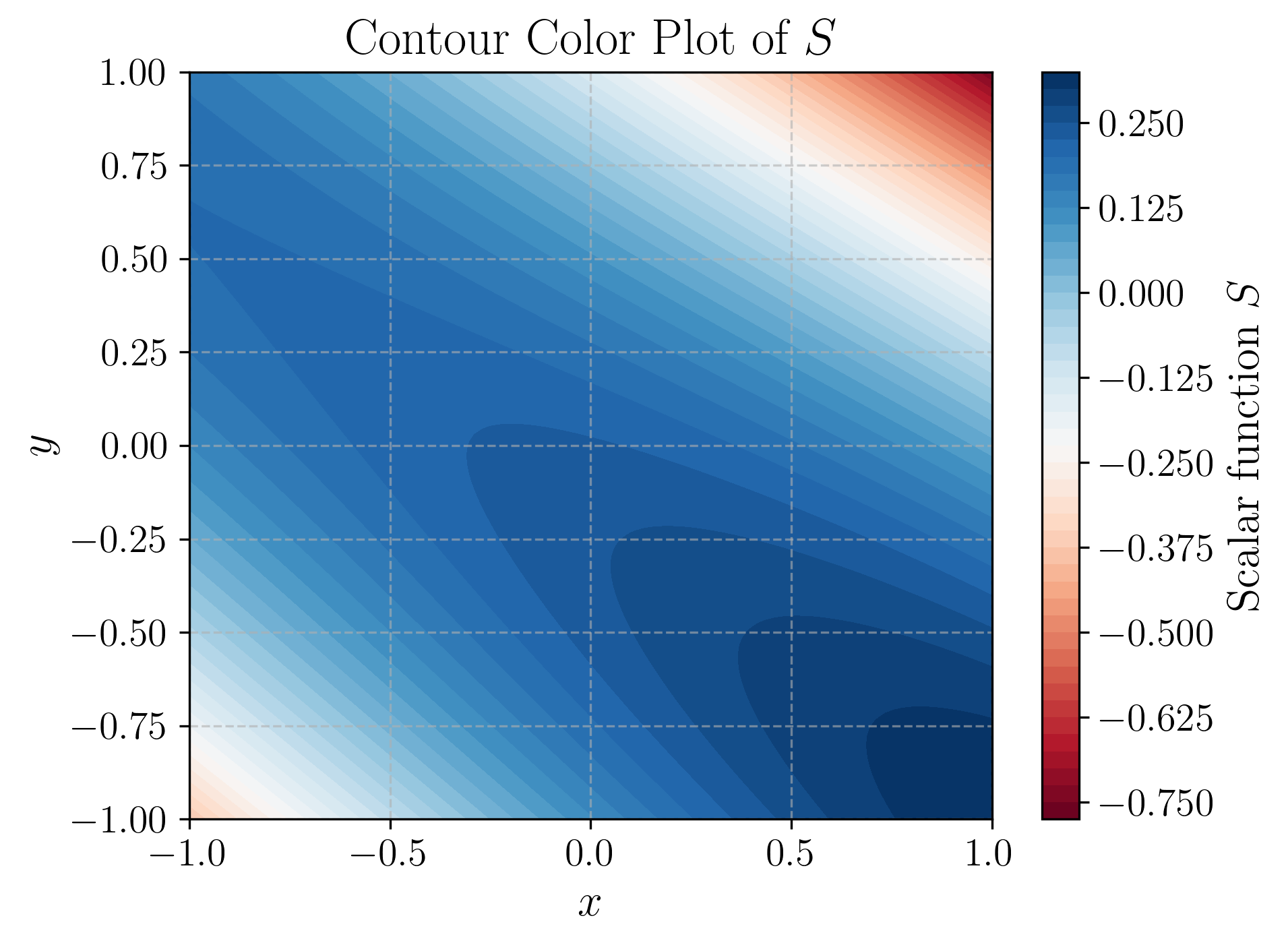}
    }
    \subfigure[2.3]{
        \includegraphics[width=0.12\textwidth]{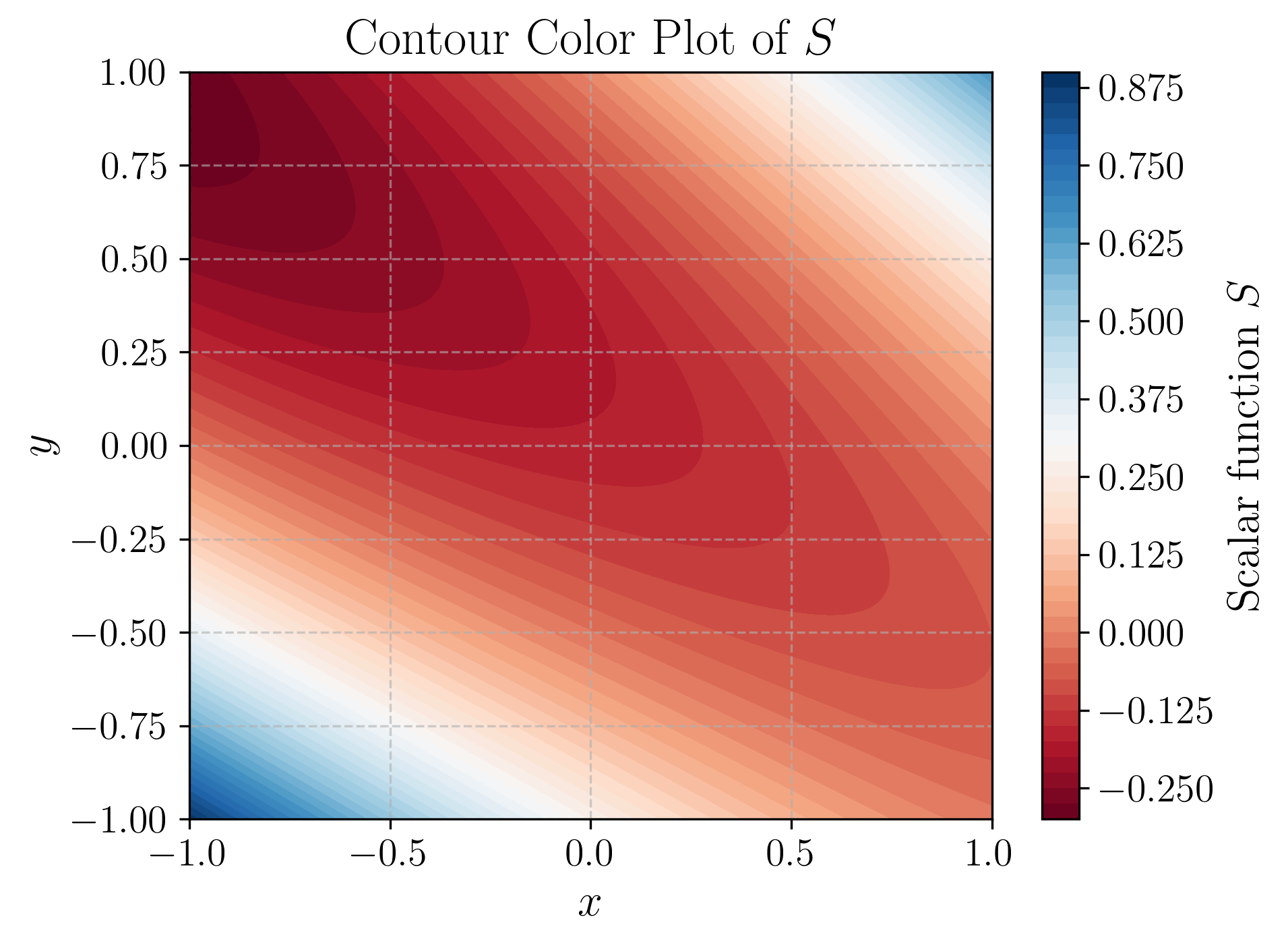}
    }
    \subfigure[3.1]{
        \includegraphics[width=0.12\textwidth]{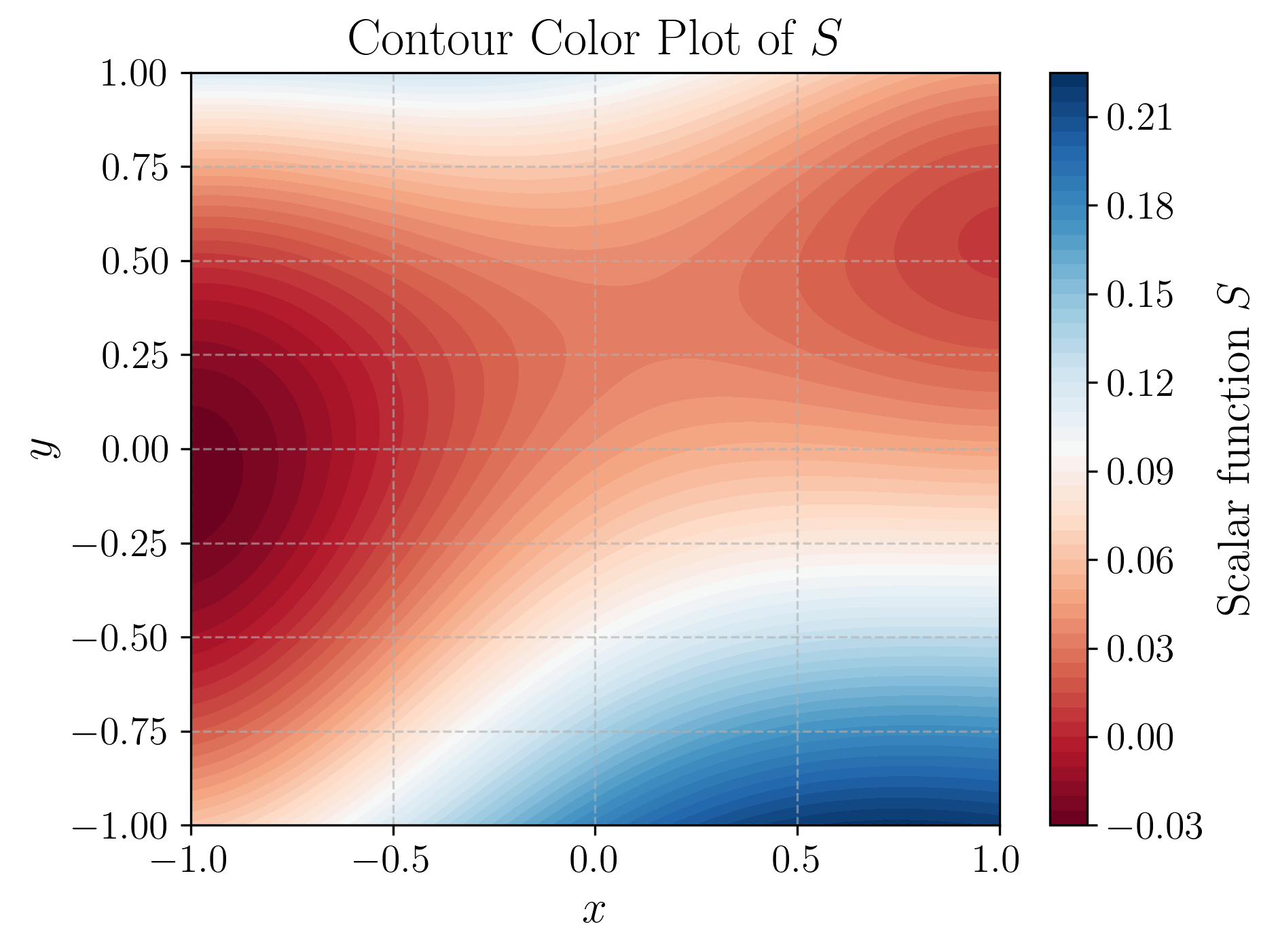}
    }
    \subfigure[3.2]{
        \includegraphics[width=0.12\textwidth]{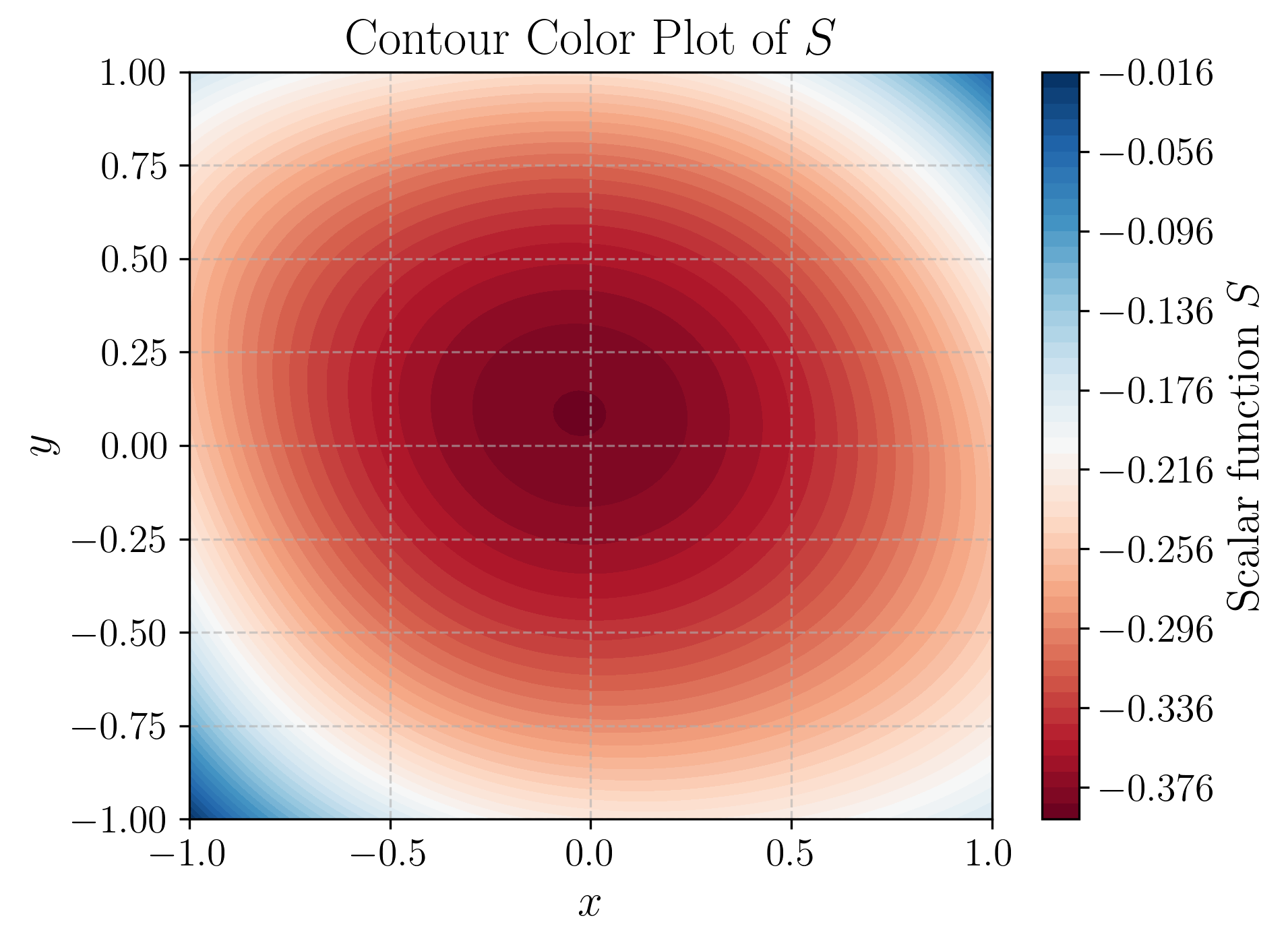}
    }
    \subfigure[3.3]{
        \includegraphics[width=0.12\textwidth]{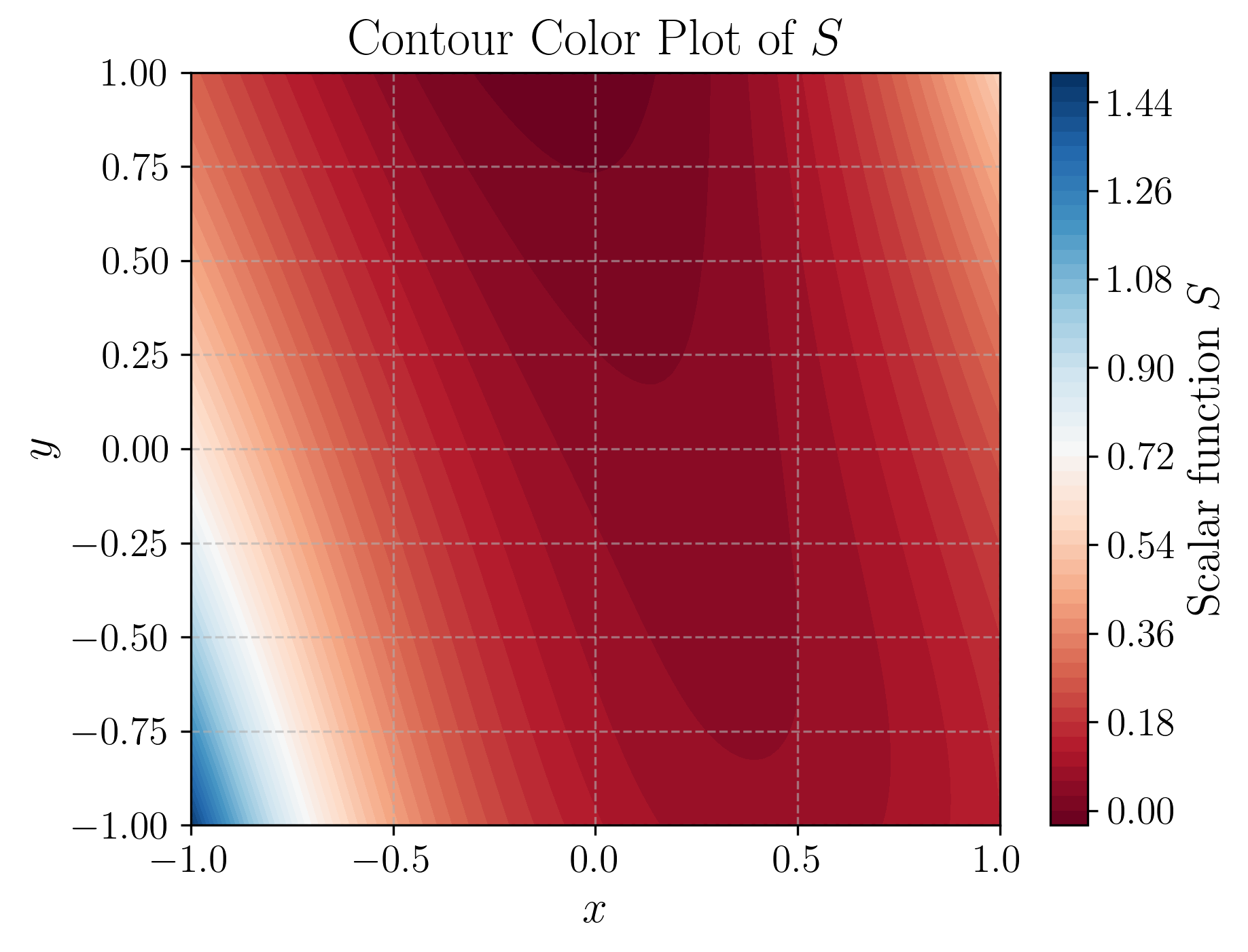}
    }
    \caption{Learned scalar functions $S$ for MASS scientist $i.j$ where $i \in \{1,2,3\}$ corresponding to SHO (top), simple pendulum(middle) and gravitational (bottom) and $j$ represents seed index.}
    \label{fig:many_contours}
\end{figure}

Below we provide some additional visualizations of the learned scalar functions $S$. Note the various shapes: elliptical, parabolic, hyperbolic, and degenerate (where the level curves are nearly straight lines). In genreral, the shape closely resembles a conic section, in large part due to the nature of these problems: kinetic and potential energy terms are typically second order in the terms of the generalized coordinates. Even for the gravitational problem where we have a $- \frac{1}{|x|}$ potential, the learnt scalar functions still resembles a conic section! 

In general, we observe that while the learned scalar function look different. The differences lie in simple parity swaps (positive to negative, elliptical to hyperbolic) and learned theories are in fact similar, according to our discussions in the main paper. The number of curves near straight lines indicate that many theories lie on the border of a ``Hamiltonian" or ``Lagrangian" contour.

\section{Higher dimensional problems}\label{sec:high_dim}
\begin{figure}[h!]
    \centering
    \includegraphics[width=\linewidth]{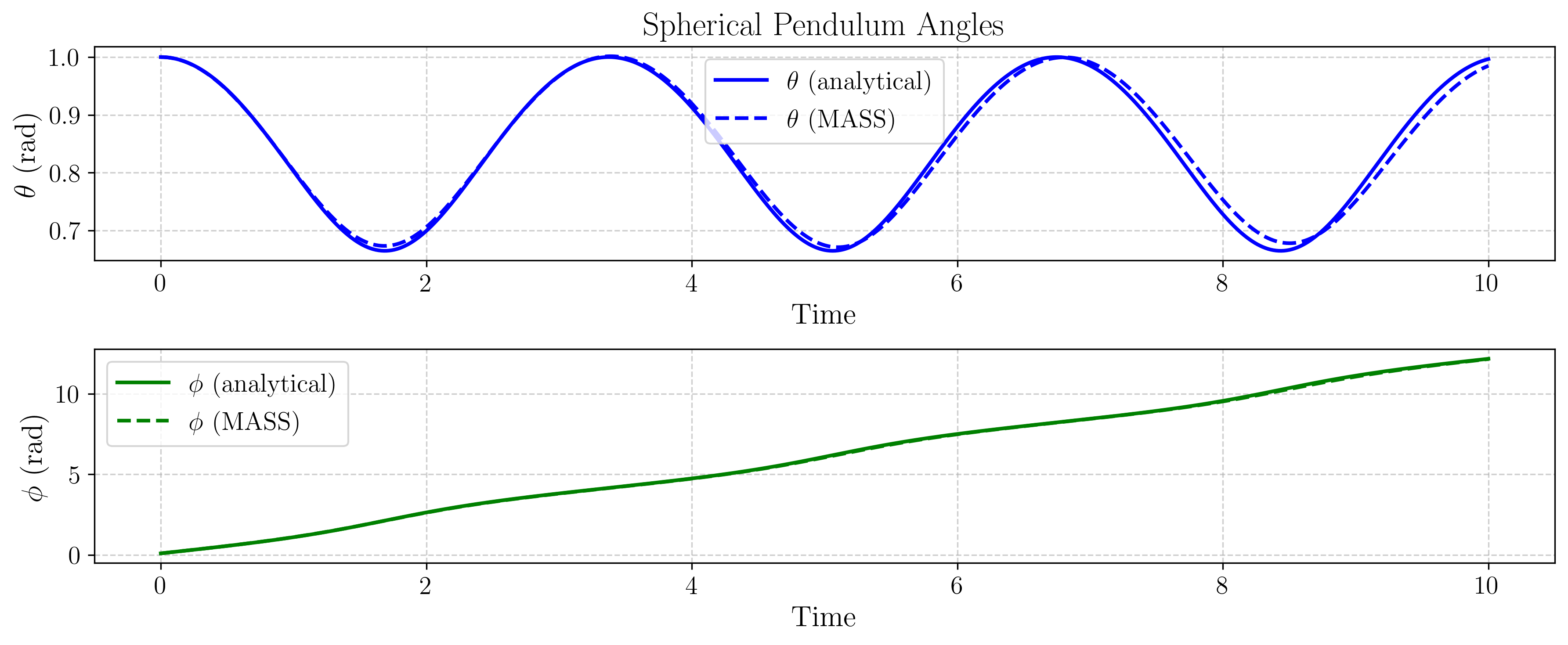}
    \caption{\label{fig:spherical_pendulum} Comparison of MASS and analytical solution to the spherical pendulum, solved with initial conditions $(\theta, \phi) = (1, 0.1), (\dot{\theta} ,\dot{\phi}) = (0, 1)$. }
\end{figure}

\begin{figure}[h!]
    \subfigure[Trajectories]{
        \includegraphics[width=0.4\textwidth]{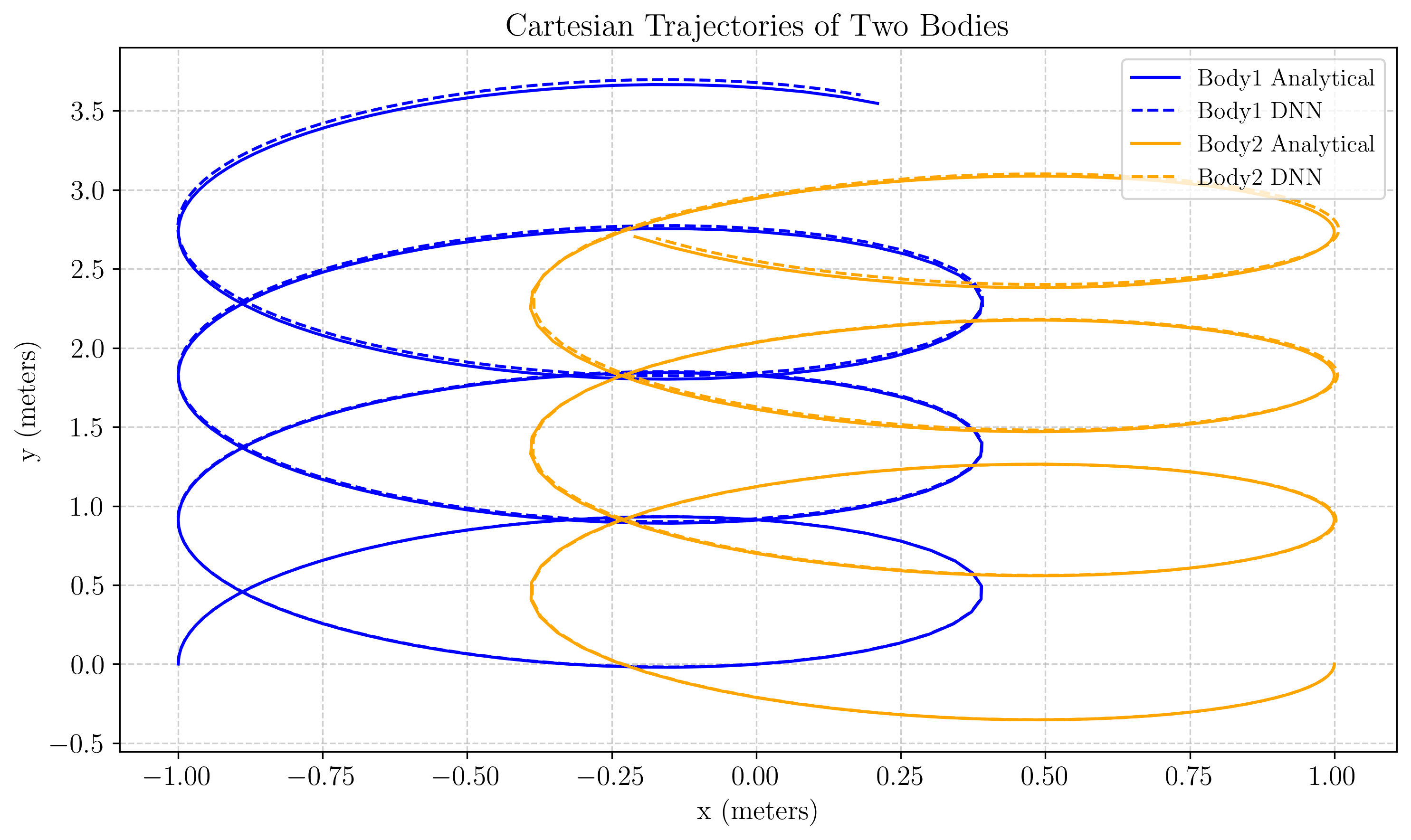}
    }
    \subfigure[Coordinates]{
        \includegraphics[width=0.4\textwidth]{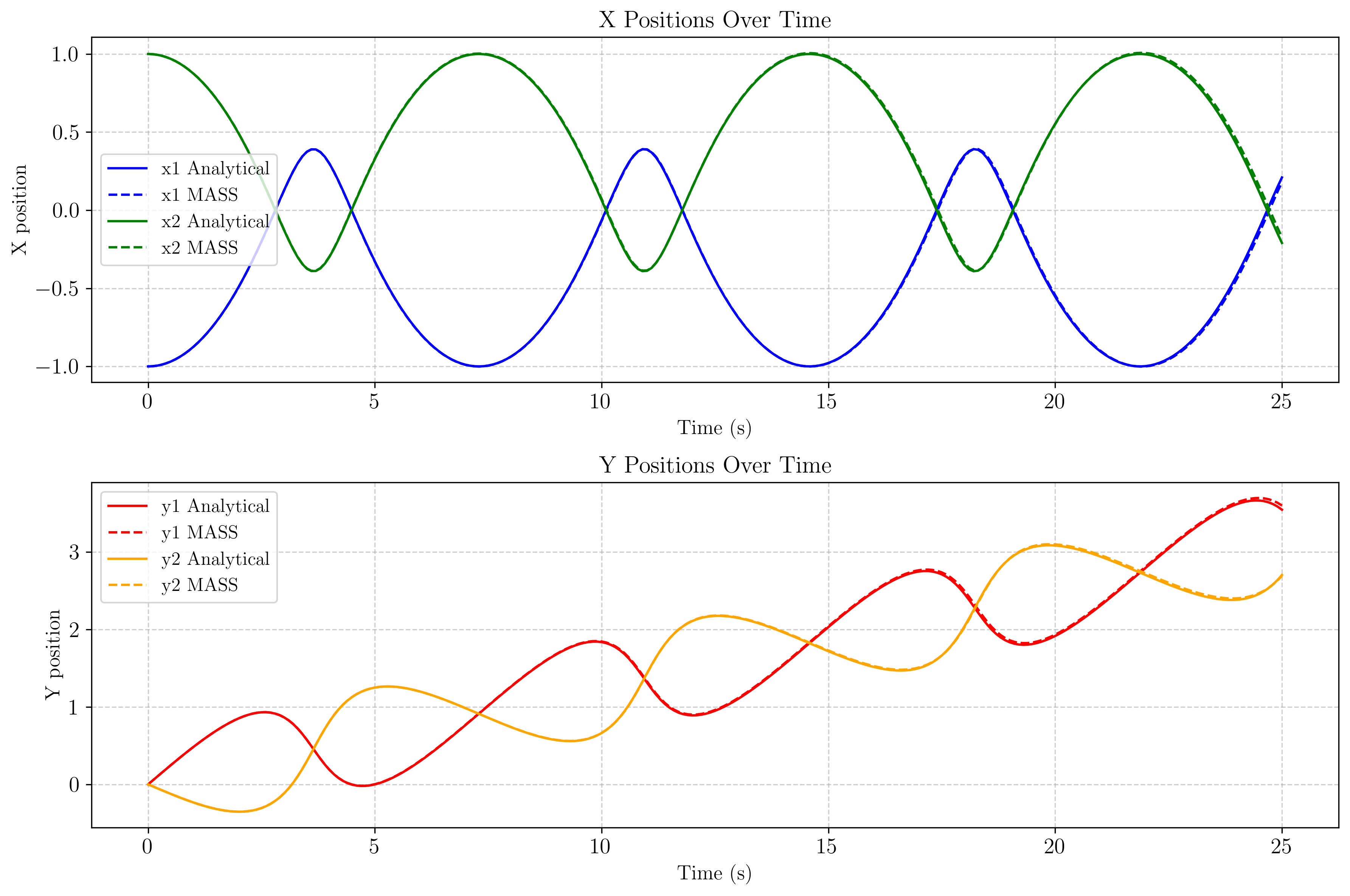}
    }
    \caption{A four-dimensional problem. Two body solution for MASS compared to the analytic solution. Problem is posed in Cartesian coordinates, with the initial conditions $(x_1, y_1, x_2, y_2) = (-1, 0, 1, 0), (\dot{x}_1, \dot{y}_1, \dot{x}_2, \dot{y}_2) = (0, 0.5, 0, -0.25)$.}
    \label{fig:two_body}
\end{figure}

\begin{figure}[ht!]
    \subfigure[Trajectories]{
        \includegraphics[width=0.4\textwidth]{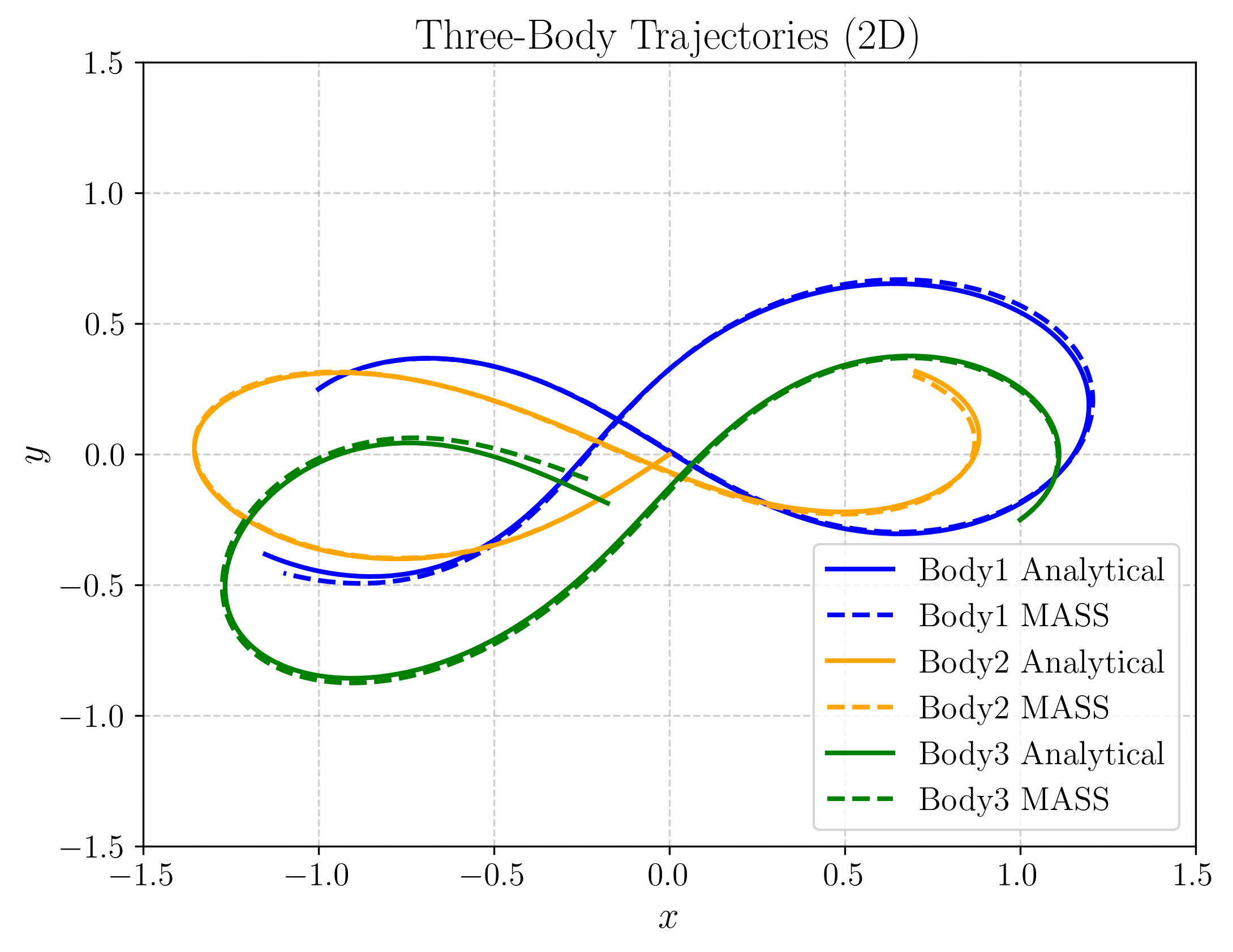}
    }
    \subfigure[Coordinates]{
        \includegraphics[width=0.4\textwidth]{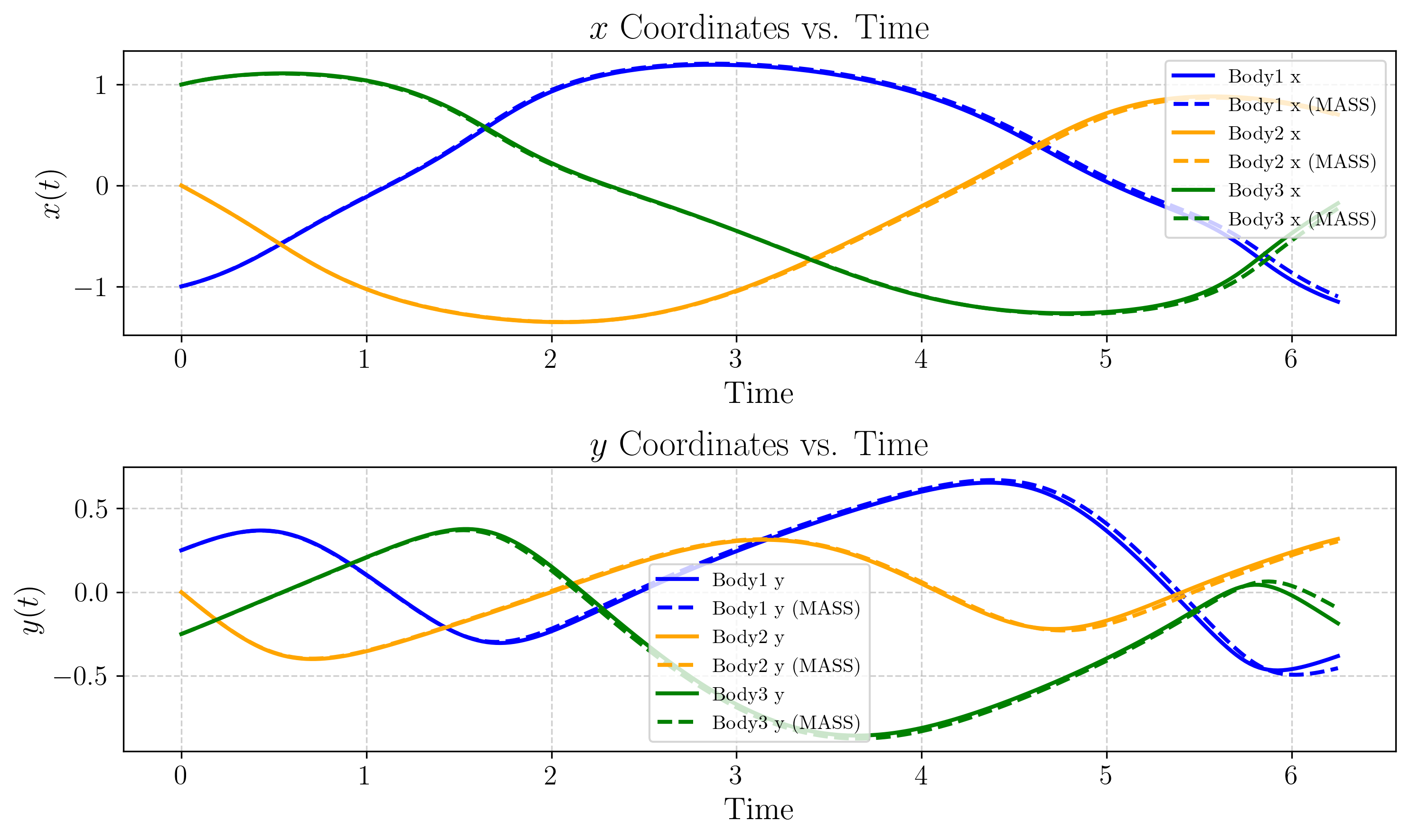}
    }
    \caption{A six-dimensional problem. Three body solution for MASS compared to the analytic solution. Problem is posed in Cartesian coordinates, with the initial conditions $(x_1, y_1, x_2, y_2, x_3, y_3) = (-1, 0.25, 0, 0, 1, -0.25), (\dot{x}_1, \dot{y}_1, \dot{x}_2, \dot{y}_2, \dot{x}_3, \dot{y}_3) = (0.45,  0.43, -1.,   -0.9,  0.44,  0.43)$. This is a slight perturbation from the stable figure-8 solution.}
    \label{fig:three_body}
\end{figure}

We can apply MASS to also solve higher dimensional problems, beyond that of the double pendulum in Figure~\ref{fig:double_pendulum}.  In particular, we demonstrate the ability of MASS to solve for periodic orbits reliably.

Another natural extension to the simple pendulum to two dimensions is the spherical pendulum, parameterized by the two degrees of freedom $\theta$ and $\phi$. We present a typical solution displaying the oscillation about an equilibrium conical solution given by $\dot{\phi} = \sqrt{\frac{g}{L \cos\theta}}$. $\dot{\theta}$ oscillates in a near harmonic motion while $\dot{\phi}$ oscillates about a constant drift which is the initial angular velocity. 

The exact equations of motion are given by 
\begin{align*}
    \ddot{\theta} &= \sin\theta \cos\theta \, \dot{\phi}^2 - \frac{g}{L} \sin\theta \\
    \ddot{\phi} &= -\frac{2 \dot{\theta} \dot{\phi} \cos\theta}{\sin\theta}
\end{align*}
with the energy of the system given by 
\begin{align*}
    E = \underbrace{\left[ \frac{1}{2} m l^2 \dot{\theta}^2 + \frac{1}{2} m l^2 \sin^2\theta \dot{\phi}^2 \right]}_{T} 
+ \underbrace{\left[ -mgl\cos\theta \right]}_{V}
\end{align*}
and we set all physical constants to $1$ for the purpose of this experiment.

Another problem we can tackle with MASS is the $n$-body problem. The $n$-body problem involves  interacting masses under the influence of gravitational forces. The equations of motion for the -body problem are given by:
\begin{align*}
m_i \ddot{\mathbf{r}}_i &= \sum_{j \neq i} G \frac{m_i m_j}{|\mathbf{r}_j - \mathbf{r}_i|^3} (\mathbf{r}_j - \mathbf{r}_i), \quad i = 1, 2, ..., n.
\end{align*}
where $m_i$ is the mass of the $i$-th body, $\mathbf{r}_i$ is its position vector, and $G$ is the gravitational constant. As with all previous problems, we set all physical constants to 1.

For the two-body problem in Cartesian coordinates, represented by $\mathbf{x} = (x_1, y_1, x_2, y_2)$, we report the comparisons between the analytic and MASS results in Figure \ref{fig:two_body}, from which we can observe an accurate learning of the behavior including the drift of the two bodies as well as their orbits about the common center of mass. Note that this problem can effectively be reduced to two dimensions with a coordinate transform using the reduced mass, but nonetheless MASS is able to learn the higher dimensional general representation in Cartesian coordinates.

The two-body problem is not so difficult. But what about the three-body problem? This is known to be chaotic. Turns out, we can solve this too! We can use MASS on this problem directly in 6 dimensions, and the result is shown in Figure~\ref{fig:three_body}. The initial conditions are chosen to be a deviation from the known stable figure-8 solution, and shows that MASS can capture the interaction between all three bodies accurately.

For all the systems presented, we use Runge-Kutta fourth-order integration solver for the ODE. Together with the accuracy of the MASS solver, the integration solver conserves the energy of the systems significantly.
Again, we are not claiming that MASS is the state-of-the-art solver for physical systems. In fact, many of these toy examples are not solved to the best precision, and some are only exhibited near equilibrium states of which the behavior of the system is regular. In fact, a persistent problem is the stability of training of MASS, which is accentuated in irregular regimes. Nonetheless, the ability of MASS to be adapted to higher dimensional problems without much change in architecture and hyperparameters is a promising sign in building general and interpretable AI physics scientists in the future.






\end{document}